%% file: CanAI2026-PMLR.tex
\documentclass[10pt]{cai26}
\usepackage{graphicx}
\usepackage{subcaption}
\usepackage{algorithm}
\usepackage{algpseudocode} % from algorithmicx
\usepackage{amsthm}
\newtheorem{corollary}{Corollary}
\usepackage{graphicx}
\usepackage{subcaption}
\usepackage{xspace}
\usepackage[font=small]{caption}
\usepackage{eso-pic}

\newcommand{\acceptednotice}{%
This is the preprint of the work accepted for publication in the Proceedings of the 39th Canadian Conference on Artificial Intelligence (Canadian AI 2026)%
}

\begin{document}
% Editorial staff will replace the following values:
% 1. Conference Year
% 2. Issue number
% % 3. Article DOI
% \def\conferenceyear{2026}
% \volumeheader{39}{0}%{00.000}

% Keep the macros defined, but suppress the volume header safely
\def\conferenceyear{2026}

\makeatletter
\@ifundefined{volumeheader}{%
  \newcommand{\volumeheader}[2]{}%
}{%
  \renewcommand{\volumeheader}[2]{}%
}
\makeatother

\volumeheader{}{}

\begin{center}

% \begin{center}
% \footnotesize
% This is the preprint of the work accepted for publication at the preceding 39th Canadian Conference on Artificial Intelligence (Canadian AI 2026).
% \end{center}

\vspace{0.5em}

\title{CEAR: Certified Ensemble Adversarial Robustness in DNNs}
\maketitle

\AddToShipoutPictureFG{%
  \AtPageUpperLeft{%
    \raisebox{-0.40in}{%
      \makebox[\paperwidth][c]{\scriptsize \acceptednotice}%
    }%
  }%
}

\thispagestyle{empty}
\pagenumbering{gobble}

% Add Authors and Affiliations in the camera ready
% for the double blind review, please leave this section as is 
\begin{tabular}{cc}
Daniel Sadig\upstairs{*\affilone}, Mohammadreza Maleki\upstairs{\affilone}, Hamed Karimi\upstairs{\affilone}, Reza Samavi\upstairs{\affilone\affiltwo}
\\[0.25ex]
{\small \upstairs{\affilone} Department of Electrical, Computer, and Biomedical Engineering,} \\
{\small Toronto Metropolitan University, Toronto, ON, Canada} \\
{\small \upstairs{\affiltwo} Vector Institute, Toronto, ON, Canada} \\
\end{tabular}
  
% Replace with corresponding author email address
\emails{
  \upstairs{*} \{daniel.sadig, mohammadreza1.maleki, hamed.karimi, samavi\}@torontomu.ca 
  % \upstairs{*} mohammadreza1.maleki@torontomu.ca
  % \upstairs{*} mohammadreza1.maleki@torontomu.ca
}
\vspace*{0.1in}
\end{center}

\begin{abstract}
Deep Neural Networks (DNNs) are highly susceptible to adversarial perturbations, leading to extensive research on robustness for safety-critical applications. State-of-the-art empirical defense mechanisms improve the robustness of DNNs through the training phase, but still struggle against adaptive white-box attacks. On the other hand, certified defenses offer provable guarantees of robustness within a specified perturbation bound. These guarantees hold regardless of the level of perturbations, even if the attacker is given full knowledge of the model. In this paper, we propose \emph{CEAR}, an ensemble-based robust method that utilizes a hybrid of empirical and certified defense mechanisms. CEAR trains each network within the ensemble using varying Gaussian noise and temperatures to obfuscate gradients and logits, making the model more resistant to stronger gradient-based attacks. We then use noisy logits and propose two different voting mechanisms to further improve robustness. Furthermore, we extend randomized smoothing to verify the robustness of ensemble-based classifiers. Our experimental evaluations on MNIST, CIFAR10, and TinyImageNet datasets demonstrate superior certified accuracy on average, increased robustness radius, and decreased transferability compared to baseline methods.

\end{abstract}

% add your keywords
\copyrightnotice

%{\noindent{\bf Editors:} Lydia Bouzar-Benlabiod, Carson K. Leung}
\input{Sections/Introduction}

\input{Sections/Method}

\input{Sections/Verification}
\input{Sections/Evaluations}

\input{Sections/Conclusion}

\section*{Acknowledgements}
This research was undertaken thanks in part to funding from the Canada First Research Excellence Fund at Toronto Metropolitan University and Natural Sciences and Engineering Research Council of Canada (NSERC) Discovery grants (\#348100).

\clearpage
\printbibliography[heading=subbibintoc]

\clearpage
\appendix
\input{Sections/Appendix}

% % All references should be stored in the file "references.bib".
% % That call to use that file is in "cai.cls". 
% % Please do not modify anything below this line.

\end{document}

%% file: Sections/Introduction.tex
\section{Introduction}
Modern Deep Neural Networks (DNNs) achieve high clean accuracy on independently and identically distributed (i.i.d.) test sets~\cite{aldahdooh2022adversarial}, but remain highly vulnerable to adversarially crafted perturbations, small and often imperceptible input modifications that can drastically alter the model predictions. The threat of these adversaries has hindered the deployment of DNNs in safety-critical tasks~\cite{szegedy2014intriguing, carlini2017towards}.
To counter these threats, researchers have proposed two broad categories of defense mechanisms. \textit{Empirical defenses}, such as adversarial training~\cite{wang2025failure} and defensive distillation~\cite{papernot2016distillation}, which modify the training process, often by incorporating adversarial examples or altering the training process to improve robustness. Nevertheless, these defense techniques do not offer a formal robustness guarantee and have been shown to fail under strong or adaptive attacks. On the other hand, \textit{certified defenses} provide formal, provable guarantees that classifiers prediction remains invariant within a specific perturbation boundary~\cite{li2023sok}. Certified defense approaches such as convex relaxation methods~\cite{raghunathan2018semidefinite} and randomized smoothing~\cite{lecuyer2019certified, cohen2019certified, zuhlke2025adversarial} all provide formal guarantees of robustness. Randomized smoothing verifies that the prediction of a smoothed classifier remains consistent within specified perturbation radii by aggregating predictions over perturbed inputs.

Despite advances in certified defense methodologies, enhancing the robustness of individual classifiers at larger perturbation radii remains a fundamental challenge. 
Recent investigations have demonstrated that ensemble-based approaches have shown improved robustness through the aggregation of diverse model predictions~\cite{yangcertified}. However, they still exhibit a steep decline in certified accuracy under larger perturbation budgets. 

In this paper, we propose CEAR, \textit{Certified Ensemble Adversarial Robustness}, a defense method that improves the robustness of DNNs via a certified ensemble approach. 
CEAR is inspired by three existing defense approaches: Gaussian augmentation~\cite{cohen2019certified, lecuyer2019certified}, distillation with temperature scaling~\cite{papernot2016distillation}, and noisy logits~\cite{liang2023advanced}. Although each of these three approaches has individually demonstrated effectiveness in improving the robustness of DNNs, their combined impact remains unexplored. 
%CEAR integrates both empirical and certified techniques across training and inference phases. 
% To utilize randomized smoothing, we train an ensemble of networks with \emph{Variable Gaussian Augmentation} (VGA), which adds Gaussian noise with distinct standard deviations to each network. VGA improves robustness and reduces adversarial transferability, the ability of adversarial examples crafted for one network to fool other networks in the ensemble~\cite{chen2024diversity}. Additionally, we apply \emph{Distillation with Temperature Scaling} (DTS) to smooth decision boundaries, reducing the success rate of gradient-based attacks. 

To leverage randomized smoothing, we train an ensemble of networks employing \emph{Variable Gaussian Augmentation} (VGA), which augments each network with Gaussian noise sampled from distinct standard deviations. VGA enhances robustness while mitigating adversarial transferability, the phenomenon whereby adversarial examples crafted for one network successfully compromise other ensemble members~\cite{chen2024diversity}. Furthermore, we incorporate \emph{Distillation with Temperature Scaling} (DTS) to smooth decision boundaries, thereby diminishing the effectiveness of gradient-based attacks. 
At inference time, we apply Gaussian noise to the input data to produce \emph{noisy logits}, making it more difficult for stronger adversarial attacks (such as C\&W~\cite{carlini2017towards}) to recover the original logits. 
Predictions across the ensemble are then aggregated via \emph{Geometric Median} (GM) and \emph{Robust Weighted Ensemble} (RW) voting mechanism, which unifies decision boundaries and improves robustness for varying confidence regimes. 
Finally, we verify the robustness of an ensemble by extending randomized smoothing to compute the robustness radius for each input in the dataset.

\vspace{0.1cm}
\noindent\textbf{Related Work.}
Empirical defenses, such as adversarial training~\citep{aldahdooh2022adversarial} and defensive distillation~\citep{papernot2016distillation}, attempt to smooth decision boundaries and reduce gradient sensitivity but remain vulnerable to strong gradient-based white-box attacks, such as C\&W~\citep{carlini2017towards} and AutoAttack~\citep{croce2020reliable}. To protect against these attacks, we use noisy logits to further obfuscate the gradient~\citep{liang2023advanced}. However, empirical defenses that rely on gradient obfuscating techniques lead to a false sense of security~\citep{athalye2018obfuscated}.
Therefore, we incorporate certified defenses such as randomized smoothing~\citep{cohen2019certified}, which have advanced robustness guarantees by optimizing sampling efficiency~\citep{zhai2020macer} and tightening robustness bounds~\citep{NEURIPS2019_3a24b25a, li2022double}.
To further improve robustness, researchers proposed training an ensemble of independently smoothed classifiers paired with adaptive ensemble voting~\citep{huang2023fasten, yangcertified, liu2021enhancing}.
Most existing ensemble methods assume a uniform noise distribution across networks, limiting the network's diversity and increasing transferability~\citep{chen2024diversity}. To overcome this limitation, we trained each network in the ensemble with a distinct noise parameter, resulting in diverse gradient distributions and preventing the success of transfer-based attacks. Furthermore, classical weighted ensemble techniques~\citep{jimenez1998dynamically, zhang2021towards} have shown limited improvements in certified accuracy for larger perturbation budgets, which we mitigate by our proposed voting mechanisms. These improvements decrease transferability and increase robustness of the ensemble-based classifier (Section~\ref{sec:experiments}).

\vspace{0.1cm}
\noindent\textbf{Contributions.} We are making the following contributions:
(1) We propose CEAR, a certified adversarial defense framework that unifies DTS and VGA in an ensemble training pipeline to promote decision-boundary smoothness and diversity. At inference, CEAR leverages noisy logits to hinder gradient-based attacks and applies two voting mechanisms to recover clean accuracy while improving robustness. (2) We extend the randomized smoothing verification method to verify the robustness of ensemble-based classifiers. 
(3) Our experimental evaluations demonstrate that CEAR achieves a higher certified accuracy on average at larger radii and is more resilient to gradient-based white-box attacks than existing state-of-the-art baselines.

\begin{figure*}[t]
  \centering
  \begin{subfigure}[t]{0.53\linewidth}
    \centering
    \includegraphics[width=\linewidth]{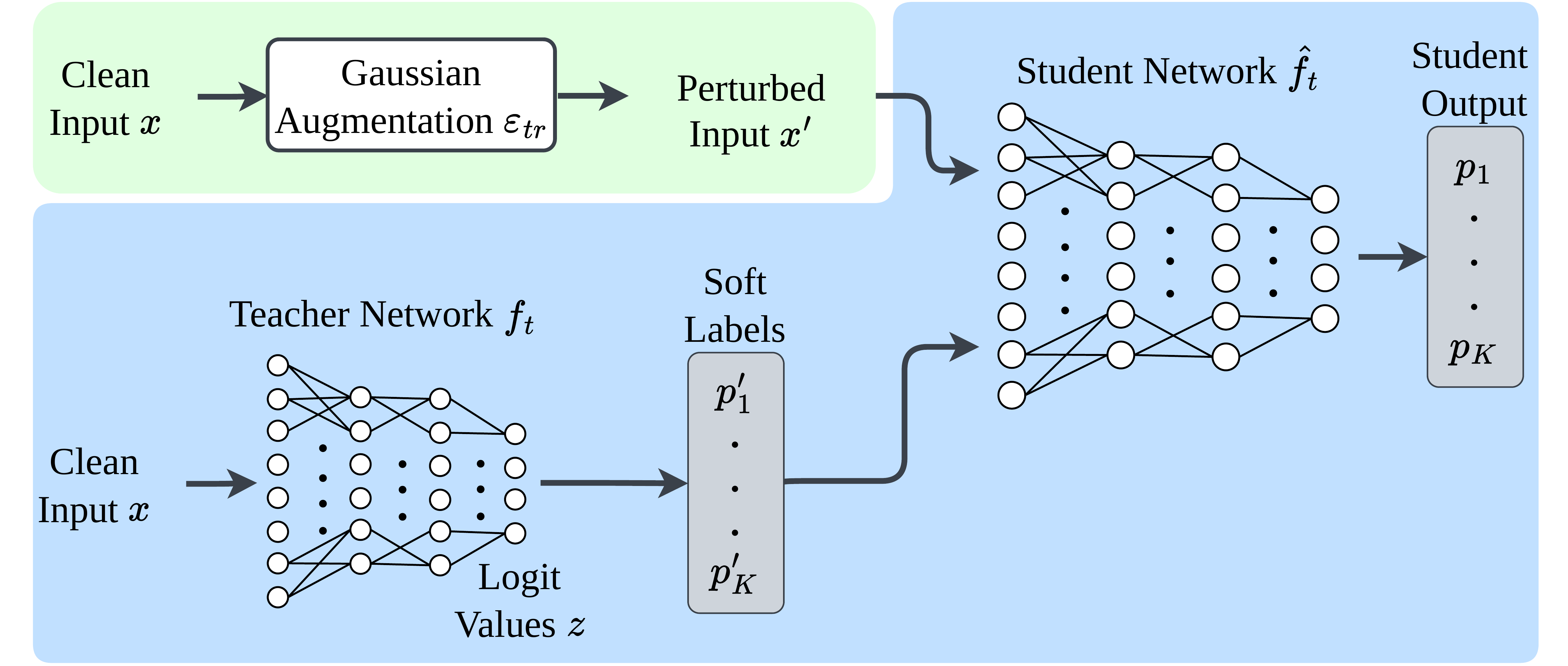}
    \caption{Training: Use clean input $x$ to train teacher network to generate soft labels; train student network using the soft labels plus perturbed input $x'$.}
    \label{fig:training}
  \end{subfigure}\hfill
  \begin{subfigure}[t]{0.44\linewidth}
    \centering
    \includegraphics[width=\linewidth]{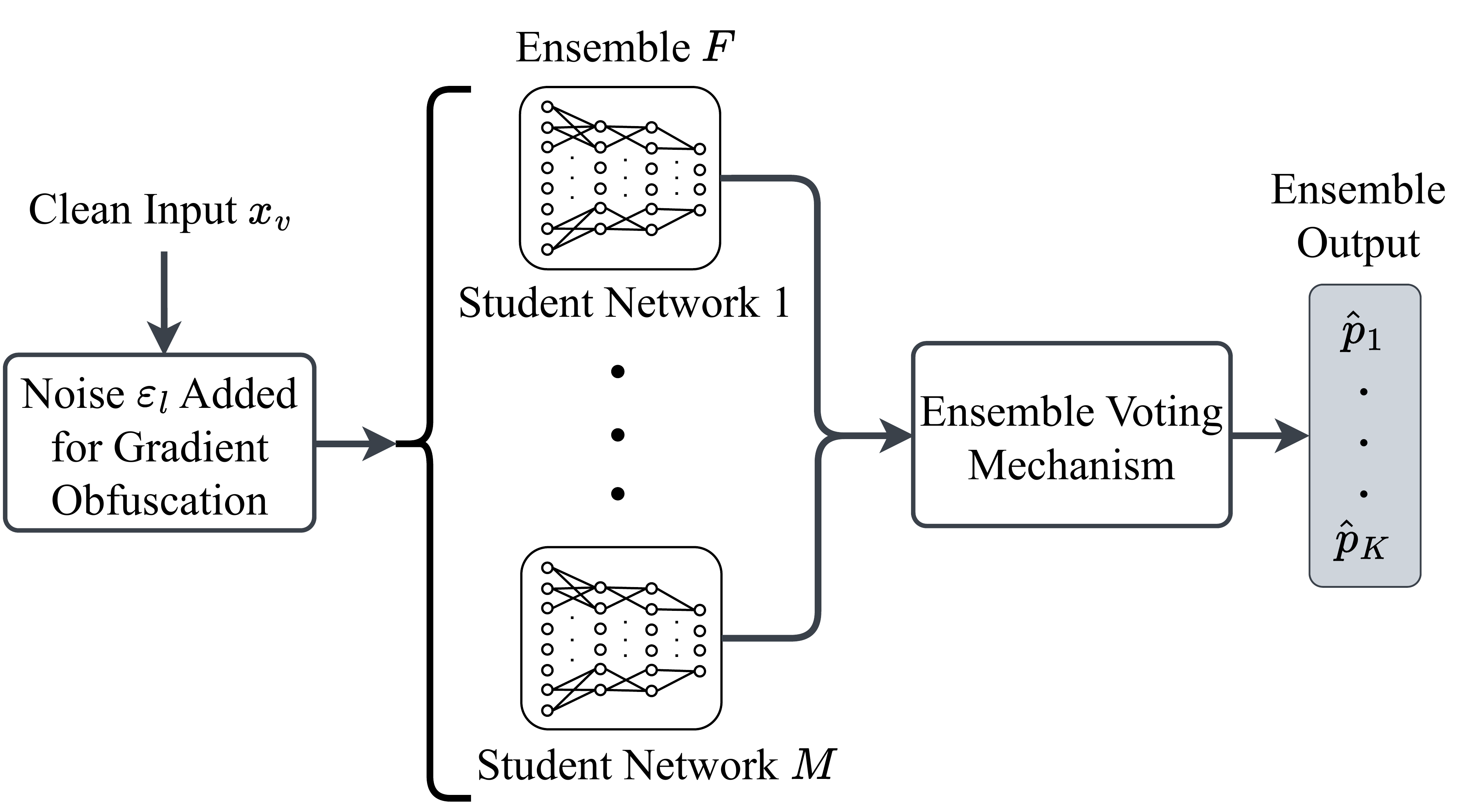}
    \caption{Inference: Add asynchronous noise to ensemble for gradient obfuscation and generate output using a voting mechanism.} 
    \label{fig:inference}
  \end{subfigure}
  \caption{
 % The training phase (a) begins by training the teacher network to generate soft labels. The input of the student network is perturbed with $\varepsilon_{tr}$. During inference (b), noise $\varepsilon_l$ is added to the validation data to produce noisy logits, and the ensemble predictions are aggregated via our ensemble voting mechanism.
  Overview of Certified Ensemble Adversarial Robustness}
  \label{fig:Ensemble}
  \vspace{-3mm}
\end{figure*}
% The remainder of the paper is organized as follows: Section~\ref{sec:CEAR} introduces our hybrid ensemble-based methodology for adversarial robustness. We detail the training phase in Section~\ref{sec:training}, the inference phase in Section~\ref{sec:inference}, and our robustness verification in Section~\ref{sec:RV}. In Section~\ref{sec:experiments}, we present experimental results comparing our approach to baseline methods on various real-world datasets. Finally, we conclude in Section~\ref{conclusion}.

%% file: Sections/Method.tex
\section{Certified Ensemble Adversarial Robustness}
\label{sec:hear}
\vspace{-2mm}
An overview of CEAR method is presented in Figure~\ref{fig:Ensemble}. In the robust training phase (shown in Figure~\ref{fig:Ensemble}(a) and described in Section~\ref{sec:training}), we adapt Gaussian augmentation, which was initially designed for a single network (green-shaded area in Figure~\ref{fig:Ensemble}(a)), and expand this process across an ensemble of networks. In this phase, we also adapt distillation with temperature scaling (blue-shaded area in Figure~\ref{fig:Ensemble}(a)) to train a teacher network and use the output of the network in conjunction with Gaussian augmented inputs to train several student networks. 
In the inference phase (shown in Figure~\ref{fig:Ensemble}(b) and described in Section~\ref{sec:inference}), we adapt noisy logits~\citep{liang2023advanced} to obfuscate the gradients across all layers of the individual student networks in the ensemble and protect the model against stronger attacks such as C\&W attack~\citep{carlini2017towards}. Since clean inputs are exposed to several stochastic perturbations (due to the addition of Gaussian noise during training and inference phases), degradation of clean accuracy is unavoidable~\citep{wang2025failure}. Thus, we practically show that using the ensemble, along with our proposed voting mechanism, can effectively compensate for the clean accuracy deficiency.

\subsection{Robust Training}
\label{sec:training}
Let $f: \mathcal{X} \to \mathcal{Y}$ denote a  DNN, where \( \mathcal{X} \subseteq \mathbb{R}^d \) is the input space with dimension \( d \), and \( \mathcal{Y} = \{1, 2, \dots, K\} \) is the output space consisting of \( K \) class labels. Given a clean input \( x \in \mathcal{X} \), an adversarial example \( x' \) is defined as
$f(x') \neq f(x)$ subject to $\|x' - x\| \leq \varepsilon$ where \( \varepsilon \) denotes the perturbation bound under a specified norm (e.g., \( \ell_2 \) or \( \ell_\infty \)).

\vspace{0.1cm}
\noindent\textbf{Variable Gaussian Augmentation (VGA).} 
To improve the robustness of a single network, Gaussian noise is added to the input, \(x' = x + \varepsilon_{\mathrm{tr}}\), where \(\varepsilon_{\mathrm{tr}} \sim \mathcal{N}(0, \sigma^2_{\mathrm{tr}} I)\) and \(\sigma_{\mathrm{tr}}\) is the standard deviation used during training (green-shaded area in Figure~\ref{fig:Ensemble}(a))~\citep{cohen2019certified}. By training on noisy inputs, the network learns smoother decision boundaries and becomes provably less sensitive to small adversarial perturbations.
In an ensemble, each network receives distinct Gaussian perturbations that are drawn from the same fixed standard deviation \(\sigma_{\mathrm{tr}}\). However, when all networks share the same noise level, transferability remains high due to overlapping decision boundaries.
To address this vulnerability, we propose VGA, in which each ensemble network is assigned a different noise level by perturbing a base standard deviation \(\sigma_{\mathrm{tr}}\). Specifically, we sample a random variable \(\beta\) from a uniform distribution as,
\begin{equation}
\beta \sim \mathrm{Uniform}\!\left(-\tfrac{1}{2}\sigma_{\mathrm{tr}} \text{ ,}\; +\tfrac{1}{2}\sigma_{\mathrm{tr}}\right).
\end{equation}
The effective standard deviation becomes \(\sigma_{\mathrm{tr}} + \beta\), and the training-time Gaussian noise is:
\begin{equation}
\varepsilon_{\mathrm{tr}} \sim \mathcal{N}\!\left(0,\; (\sigma_{\mathrm{tr}} + \beta)^2 I\right).
\label{noise}
\end{equation}
By drawing perturbations from the distribution described in~\eqref{noise}, we generate diverse noise distributions in the ensemble and increase the diversity of the ensemble, resulting in decreasing adversarial transferability across all networks~\citep{yazdani2024denl}.
We also provide empirical evidence in Section~\ref{sec:experiments} that drawing variable Gaussian noise per input improves certified accuracy. This improvement is especially notable under large perturbation magnitudes.

\vspace{0.1cm}
\noindent\textbf{Distillation with Temperature Scaling (DTS).}
% ~\citep{papernot2016distillation} 
As shown in the blue-shaded region of Figure~\ref{fig:Ensemble}(a), 
we first train a teacher network $f_t$ to generate the predictive probability vector $\mathbb{P}\big(f_t(x)\big)$ for each input $x \in \mathcal{X}$ as, 
\begin{equation}
    \mathbb{P}\big(f_t(x)=c\big) = \operatorname{softmax}\Big( \frac{z_c}{t}\Big)\ ,
\label{softmax_eq}
\end{equation}
where $z_c$ denotes the logit value associated with the class $c\in \mathcal{Y}$, and $t \in \mathbb{R}^{[1,+\infty)}$ is a temperature parameter that smoothens the softmax distribution, resulting in softened class probabilities known as \textit{soft labels}~\citep{guo2017calibration}. 
Then, the soft labels are used as ground truth to train a student network on the Gaussian augmented input $x'$ via the loss function $\mathcal{L}$ as,  
\begin{equation}
\mathcal{L}(x') = \mathrm{CE}\Big(\mathbb{P}\big(\hat{f}(x')\big),\mathbb{P}\big(f_t(x)\big)\Big)\ , 
\end{equation}
where \(\mathrm{CE}(\cdot)\) is cross-entropy function and \( \hat{f}(x') \) denotes the logit vector associated with the student network outputs on the perturbed input $x'$. 
The advantage of using soft labels as training labels, comes from the valuable information they contain compared to the correct class. Instead of solely indicating the correct class, soft probability distributions capture the relative similarities between classes. Furthermore, utilizing DTS effectively smoothens the decision boundaries, making the transitions between different class regions in the input space more gradual. Therefore, DTS improves generalization and robustness of DNNs against unforeseen adversarial examples~\citep{papernot2016distillation, papernot2016limitations}. 

In temperature-scaled distillation, we adjust the temperature within a modest range (typically $1$ to $5$) to foster diversity across the ensemble while preserving informative gradients. Moderate temperatures allow the student to learn inter-class information without flattening the distribution.  According to~\eqref{softmax_eq}, if \(t\) is too low (\(t\approx1\)), \(\mathbb{P}\big(f_t(x)\big)\) approaches the true label, losing significance of class similarities. If \(t\) is too high, the distribution becomes nearly uniform (i.e., \(\mathbb{P}\big(f_t(x)\big)\approx1/K\)), thus flattening away informative gradients conveyed.

% \vspace{-1mm}
\subsection{Robust Inference}
\label{sec:inference}
\noindent\textbf{Noisy Logits. }The C\&W attack~\cite{carlini2017towards} revealed that defensive distillation can be circumvented when adversaries have iterative access to model logits. Building on this insight, CEAR adopts input-level randomization at inference time by injecting Gaussian noise $\epsilon \sim \mathcal{N}(0,\sigma^2 I)$ into the input rather than perturbing the logits directly~\cite{liang2023advanced}. While logit-level noise can be averaged out through repeated querying, input-level noise induces inherently stochastic logits without modifying model parameters, making adversarial reconstruction substantially harder. Thus, in CEAR the prediction $\hat{y}_x$ is obtained by applying the network layers to the perturbed input $x'$ as $\hat{y}_x = f_i \circ f_{i+1} \circ \dots \circ f_l(x')$, where $f_i$ denotes the intermediary logits and $\circ$ denotes the composition of the logits.
This stochastic forward process enhances robustness to adaptive attacks, albeit at the cost of reduced predictive confidence.

\vspace{0.1cm}
\noindent\textbf{Geometric Median Ensemble (GM).}
The geometric median is a classical robust estimator of multivariate location.
Unlike the arithmetic mean, it is insensitive to a minority of extreme inputs, ensuring that a small number of corrupted or adversarially perturbed predictions cannot dominate the aggregate.
In the ensemble setting, this implies that the aggregate prediction remains close to the majority consensus even when several student networks produce distorted outputs.

Let ${p}_{i}(x)=\mathbb{P}\big(\hat{f}_{i}(x+\varepsilon_l)\big)\in\Delta^K$ denote the $K$-class probability vector produced by the $i$-th network for input $x$. We aggregate the ensemble predictions using the geometric median of the networks' softmax outputs defined as,
\begin{equation}
\hat{{q}}(x)
\;=\;
\arg\min_{{q}\in\Delta^K}
\sum_{i=1}^M \bigl\|{q}-{p}_{i}(x)\bigr\|_2\ ,
\label{}
\end{equation}
where the optimization is constrained to the probability simplex $\Delta^K$ to ensure a valid predictive distribution.
This estimator corresponds to the GM, a classical notion of multivariate location \cite{haldane1948note}.
This optimization finds the point $\hat{q}(x)$ whose total Euclidean distance to all member probabilities $\{{p}_{i}(x)\}$ is minimal.  The final class decision is $c_A=\arg\max_c\hat q_c(x)$.
In practice, the geometric median can be efficiently computed using iterative procedures such as Weiszfeld’s algorithm \cite{weiszfeld1937point}.
Importantly, aggregation is performed at the level of softmax probabilities rather than logits or labels, preserving the probabilistic interpretation of each model’s output.

\vspace{0.1cm}
\noindent\textbf{Robust Weighted Ensemble (RW)}. To improve robustness at larger perturbation budgets while compensating for accuracy degradation, we propose RW that accounts for the relative contribution of each student network by assigning greater influence to networks with higher certified accuracy.
This approach effectively unifies the decision regions induced by the individual ensemble members.
Unlike the distinct training set \( \mathcal{X}_i \) used for each student network \( \hat{f}_i \) in the ensemble, a shared validation set \( \mathcal{X}_v \) is used for all networks where \( \mathcal{X}_i \cap \mathcal{X}_v = \emptyset \) for all \( i \in \mathbb{N}^{[1,M]} \).
In the ensemble \( F = \{\hat{f}_i\}_{i=1}^M \) of \( M \) student networks, we first perturb the input \( x_v \in \mathcal{X}_v \) with the Gaussian noise of a fixed standard deviation, resulting in perturbed inputs.
Then, we pass the perturbed validation data to each individual student network $\hat{f}_i$ separately to compute its corresponding certified accuracy $a_i$.
Finally, we compute the ensemble's predictive probability vector \( \hat{p} \) using the weighted average of the individual network probability vector \( p_i \) as,
\begin{equation}
    \hat{p}(x) = \sum_{i=1}^{M} \lambda_i p_i(x) \qquad \text{such that} \qquad \lambda_i = \frac{a_i}{\sum\limits_{j=1}^{M} a_j}\ ,
\label{weight_eq}
\end{equation}
where $\lambda_i \in \mathbb{R}^{[0,1]}$ denotes the \emph{network contribution factor} proportional to $a_i$ of each $\hat{f}_i$.
Therefore, networks with higher certified accuracy within the given perturbation level have a stronger vote in the final predictive probability $\hat{p}$ in~\eqref{weight_eq}.
By aggregating the networks' predictions in the ensemble with our robust weighted ensemble, the robustness of DNNs is improved.

Although both aggregation schemes improve robustness, they are effective in different confidence regions as validated in Section~\ref{sec:experiments}. When individual student networks produce confident and consistent predictions, we employ GM aggregation, which sharpens the ensemble decision and yields higher certified accuracy at small radii. In contrast, as the perturbation budget increases and the model confidence degrades, we adopt RW, which explicitly down-weights less reliable networks based on their certified accuracy.

\vspace{0.1cm}
\noindent\textbf{Computational Overhead.}\label{sec:complexity}
As an ensemble-based defense, CEAR increases runtime and memory by requiring $M$ student networks, but this deliberate overhead is often acceptable in safety-critical deployments where improved robustness and verifiable guarantees outweigh additional compute.
When analyzing CEAR's computational complexity, we suppress constant per-sample costs (e.g., those for forward/backward passes or logit perturbation in a single DNN) since they remain uniform across method components, and asymptotic analysis inherently ignores such implementation-level constants. Instead, we focus on how runtime grows with the main input variables in CEAR: the ensemble size \(M\), training dataset size \(\lvert\mathcal{X}\rvert\), validation dataset size \(\lvert\mathcal{X}_v\rvert\), and test sample size \(\lvert\mathcal{X}_{\text{test}}\rvert\).

During training, CEAR independently trains each of the \(M\) networks in the ensemble using DTS (teacher and student forward-backward passes as described in Section~\ref{sec:training}), and VGA is applied on‑the‑fly. 
As VGA involves generating input-dependent noise $\varepsilon_{tr}$ per sample via Gaussian sampling, this noise augmentation introduces a negligible constant per-sample overhead, i.e., $\mathcal{O}(1)$; thus, the total time to train all networks on dataset \(\mathcal{X}\) scales as
$\mathcal{O}\bigl(M . |\mathcal{X}|\bigr)$.
This reflects linear growth in both ensemble and dataset size.

After training, each network’s contribution factor is computed once on a held-out validation set $\mathcal{X}_{v}$ based on its certified accuracy in~\eqref{weight_eq}, incurring a one-time cost of $\mathcal{O}(M.|\mathcal{X}_{v}|)$. These weights are then fixed and reused during inference. For each test sample, CEAR performs noisy forward passes through the $M$ pretrained students and applies RW voting via a single linear scan over the ensemble, adding only $\mathcal{O}(M)$ overhead. Thus, per-sample inference remains $\mathcal{O}(M)$ and scales as $\mathcal{O}(M.|\mathcal{X}_{\text{test}}|)$ over the full test set.  
For CEAR with geometric median aggregation, the same $M$ noisy forward passes are used, but logits are combined via element-wise log-probability averaging across $K$ classes before prediction, incurring an additional $\mathcal{O}(M.K)$ aggregation cost. Nevertheless, the overall inference complexity remains linear in $M$, preserving CEAR’s computational efficiency.

%% file: Sections/Verification.tex
\section{CEAR Robustness Verification}
\label{sec:RV}
Prior to deployment, robust verification is essential to guarantee that a model's decisions are invariant to perturbations within a given budget.
We adapt randomized smoothing to assess the robustness of an ensemble model. Specifically, we use the \emph{robustness radius} as the maximum perturbation under a given $\ell_p$-norm that the model can tolerate without changing its prediction for a given input. A larger radius means that the model's decision remains invariant under larger noise perturbations. 
\begin{algorithm}[t]
\caption{Ensemble-based Robustness Radius Measurement}
\label{algo_certify}
\begin{algorithmic}[1]
\Require Clean input $x$; ensemble $F=\{\hat{f}_i\}_{i=1}^M$ of student networks; $\lambda_i$: The contribution factor for the $i^{th}$ student network; $N_x$: The number of perturbations of $x$
for ensemble prediction; $N$: Number of perturbations of $x$ for Monte-Carlo sampling; $\sigma_v$: Verification standard deviation; $\alpha$: Abstinence threshold;
\Ensure Predicted class $c_A$ with certified radius $R$, or abstain ($-1$)

\State $c_0 \gets [\ ]$ \Comment{Empty list of predictions}

\For{$j \gets 1$ \textbf{to} $N_x$}
  \State $\varepsilon_v \sim \mathcal{N}(0,\sigma_v^2 I)$
  \Statex \hspace{\algorithmicindent} $\displaystyle s_c \gets \sum_{i=1}^{M} \lambda_i \,
  \mathbb{P}\!\left(\hat{f}_i(x+\varepsilon_v)=c\right)\quad \forall c\in\mathcal{Y}$
  \State $c \gets \arg\max_{c\in\mathcal{Y}} \; s_c$
  \State $\textsc{Append}(c_0, c)$ \Comment{Ensemble prediction (Eq.~\ref{smooth_weight_ensemble_eq})}
\EndFor

\State $c_A \gets \textsc{Mode}(c_0)$ \Comment{Most frequent prediction}

\State $P_A \gets \textsc{SampleWithNoise}(F, x, c_A, N, \sigma_v)$
\State $\underline{P_A} \gets \textsc{LowerConfBound}(P_A\!\cdot\!N, N, 1-\alpha)$

\If{$\underline{P_A} > \tfrac{1}{2}$}
  \State \Return $(c_A, R)$ \Comment{Return class and radius (Eq.~\ref{radius_eq})}
\EndIf
\State \Return $-1$ \Comment{Abstain}
\end{algorithmic}
\end{algorithm}
We establish the robustness radius by first defining a smoothed classifier \( g: \mathcal{X} \to \mathcal{Y} \), formed by adding the Gaussian noise $\varepsilon_v$ to the inputs of base classifier \( f: \mathcal{X} \to \mathcal{Y} \) to generate a set of perturbed instances of each validation input $x \in \mathcal{X}_v$. 
For $N_x$ perturbed inputs \( x + \varepsilon_v \), the smoothed classifier $g$ returns the class with the highest probability that is obtained by: 
\begin{equation}
    g(x) = \arg\max_{c\in \mathcal{Y}}\ \mathbb{P}\big(f(x + \varepsilon_v) = c\big)\ ,
    \label{smooth_eq}
\end{equation}
where $c \in \mathcal{Y}$ denotes the class labels.
We then extend~\eqref{smooth_eq} to smooth ensemble-based classifiers under $\ell_2$-norm.
Let $\varepsilon_v \sim \mathcal{N}(0,\sigma_v^2I)$ be the Gaussian noise and $F: \mathcal{X} \rightarrow \mathcal{Y}$ be a function of an ensemble that contains a set of $M$ base networks \( \{f_i\}_{i=1}^M \) as, 
\begin{align}
F(x) = \arg\max_{c\in \mathcal{Y}}\ \sum_{i=1}^{M} \lambda_i\cdot\mathbb{P}\big(f_i(x) = c\big)\ ,
\label{smooth_weight_ensemble_eq}
\end{align}
where $\lambda_i$ denotes the contribution factor associated with each network $f_i$ in the ensemble.
We apply randomized smoothing to each individual network \( f_i \), resulting in a set of smoothed classifiers constituting the smoothed ensemble \( G \) on the set of $N_x$ perturbed inputs as, 
\begin{align}
G(x) = \arg\max_{c\in \mathcal{Y}}\ \mathbb{P}\big(F(x+ \varepsilon_v) = c\big)\ .
\label{smooth_ensemble_eq} 
\end{align} 
Since it is infeasible to \emph{exactly} compute the smoothed prediction $G(x)$ and verify its robustness, we employ Monte Carlo sampling algorithm to ensure reliable estimates of the predicted class and robustness radius with high probability~\citep{cohen2019certified}. We compute the robustness radius for the smoothed ensemble $G$ as described in the following corollary.
\vspace{-1mm}
\begin{corollary}
\label{coro1}
Given the most probable class $c_A \in \mathcal{Y}$ with the lower bound probability of $\underline{P_A}$ and the upper bound probability of the runner-up class $\overline{P_B}$ such that $\underline{P_A}$, $\overline{P_B} \in \mathbb{R}^{[0,1]}$, the following conditions are satisfied:
\begin{equation}
 \mathbb{P}\big(F(x + \varepsilon_v) = c_A\big)\ \geq\ \underline{P_A}\ \geq\ \overline{P_B} \geq\ \max_{c \neq c_A} \mathbb{P}\big(F(x + \varepsilon_v) = c\big)
\label{eq:sample3}
\end{equation}
such that $\overline{P_B} = 1 - \underline{P_A}$. Then, the ensemble $F$ is robust at $x$ within the radius $R$ if 
\begin{equation}
G(x+\delta) = c_A \qquad \forall\ \lVert \delta \rVert_2 \leq R\ ,
\label{eq:sample4}
\end{equation}
where
\begin{equation}
R = \frac{\sigma_v}{2} \Big(\Phi^{-1}(\underline{P_A}) - \Phi^{-1}(\overline{P_B})\Big)\ ,
\label{radius_eq}
\end{equation}
in which $\Phi^{-1}$ denotes the inverse standard Gaussian CDF.
\label{theorem_1}
\end{corollary}

\noindent The proof of Corollary~\ref{coro1} follows that of~\cite{cohen2019certified}; however, the base classifier is an ensemble rather than a single network. Corollary~\ref{coro1} shows that the robustness radius $R$ increases with larger verification noise $\sigma_v$ and higher confidence in the top class $c_A$, and diverges as $\underline{P_A} \to 1$. Since the Gaussian distribution has full support on $\mathcal{X}_v$, the condition $\mathbb{P}(F(x+\varepsilon_v)=c_A)=1$ implies that $F(\cdot)=c_A$ almost everywhere in the input space.

Algorithm~\ref{algo_certify} shows how to measure the robustness radius for a given input $x \in \mathcal{X}_v$. 
We first apply randomized smoothing to each base classifier $f$ in the ensemble \(F\) by convolving $f$ with $\varepsilon_v$. 
We generate a set of $N_x$ perturbed instances of $x$ by adding $\varepsilon_v$ (Line $3$) and collecting all the votes of the smoothed ensemble $G$ in $c_0$ (Line $4$). 
The ensemble prediction $c_A$ is then obtained by selecting the class prediction that the ensemble is most likely to predict under the given noise $\varepsilon_v$ (Line $7$). We then compute the probability \( P_A \) that the ensemble predicts the class \( c_A \) across $N$ different perturbed instances of $x$ drawn from the same noise level $\varepsilon_v$, using the \emph{SampleWithNoise} procedure (Line $8$).
Next, we compute a lower confidence bound on \( P_A \) using the \emph{LowerConfBound} procedure, which estimates the minimum probability that the ensemble predicts the class \( c_A \) under noise level \( \varepsilon_v \) (Line $9$). Here, \( 1 - \alpha \) denotes the confidence level, e.g., a 95\% confidence corresponds to \( \alpha = 0.05 \). The model is certifiably robust at $x$ if this lower bound $\underline{P_A}$ exceeds the threshold $0.5$ (Line $10$). If the bound is greater than $0.5$, we return \( c_A \) along with a certified robustness radius \( R \), within which the prediction remains provably unchanged (Line $11$). Otherwise, the ensemble model abstains, indicating insufficient confidence for certification (Line $13$).

%% file: Sections/Evaluations.tex
\section{Experimental Evaluation}
\label{sec:experiments}

We evaluated the robustness of CEAR under two voting mechanisms: GM voting denoted by CEAR(GM), and RW voting denoted by CEAR(RW), through three different experiments: (1) examine the certified accuracy for varying radii, (2) compute the robustness radius, and (3) determine which networks are less susceptible to AutoAttack under $\ell_2$ and $\ell_{\infty}$ norms. We further extended our evaluations by examining the ablated variant of CEAR without the VGA, denoted as $\mathrm{CEAR}^{-}$.

\vspace{1mm}
\noindent\textbf{Datasets and Baselines.} We evaluated CEAR against state-of-the-art baselines on MNIST \cite{LeCun_Burges_Cortes_2010}, CIFAR10~\cite{Krizhevsky_2012} and TinyImageNet~\cite{deng2009imagenet} datasets. These datasets provide a diverse evaluation setting, ranging from low-dimensional grayscale digits to high-resolution natural images with a large number of classes. The baselines are denoted by \textit{RandSmooth} for the randomized smoothing method using a single network~\cite{cohen2019certified}, and \textit{SWEEN} for the smooth weighted ensemble method~\cite{liu2021enhancing}. 

\vspace{1mm}
\noindent\textbf{Threat Models.} 
We evaluated robustness under white-box attacks, where the adversaries have full access to each network's architecture, weights, and gradients. We used a subset of \textit{AutoAttack}~\citep{croce2020reliable} which is a parameter-free and reliable evaluation framework comprised of three attacks, APGD-CE, APGD-DLR, and FAB. These attacks effectively probe decision boundaries and detect gradient
masking. Making it an effective tool to evaluate adversarial transferability across ensemble configurations.

\vspace{1mm}
\noindent\textbf{Implementation Details.}
We implemented the proposed method\footnote{The source codes are available at \url{https://github.com/tailabTMU/CEAR}.}
% \footnote{The source codes are provided in the supplementary materials.} 
using the TensorFlow framework \citep{abadi2016tensorflow} in Python
and conducted the experiments using a Tesla T4 GPU. 
For consistent comparative evaluations on MNIST and CIFAR10 with baseline methods, we used three ensembles, each consisting of five student networks individually trained with soft labels produced by teacher networks at temperatures \( t = \{2, 3, 4, 5, 6\} \). The student networks were trained and tested under varying Gaussian noise levels \( \{0.25,\ 0.5,\ 1.0\} \) for CIFAR10 and MNIST, and \( \{0.125,\ 0.25\} \) for TinyImageNet. The use of lower-noise variants on TinyImageNet is motivated by the dataset's lower test accuracy, which makes the classification task significantly more challenging and renders the model highly sensitive to larger noise perturbations.
For MNIST, we used LeNet5~\citep{lecun2002gradient} to train both the teacher and student networks using SGD Optimizer and Gaussian noise $\varepsilon_l=0.3$. % for $90$ epochs. % Table~\ref{MNIST_setup} 
For CIFAR10 and TinyImageNet, we used the same hyperparameters with Gaussian noise $\varepsilon_l=0.03$ on ResNet110~\citep{he2016deep} and ResNet18 models, respectively. Furthermore, on 10,000 CIFAR10 instances, RandSmooth takes 33(s), SWEEN takes 164(s), CEAR(RW) takes 162(s), and CEAR(GM) takes 171(s), while on MNIST, RandSmooth takes 2(s), SWEEN takes 10(s), CEAR(RW) takes 7(s), and CEAR(GM) takes 14(s).

\begin{table*}[t]
\centering
\caption{Certified accuracy (\%) at varying radii and $\sigma_v$ on MNIST (M) and CIFAR10 (C).}
\vspace{-2mm}
\resizebox{\textwidth}{!}{%
\begin{tabular}{c c cc cc cc cc cc cc cc cc cc}
\toprule
% \multirow{1}{*}{\textbf{Radius $R$}} 
\multicolumn{2}{c}{\textbf{Radius $R$}}
& \multicolumn{2}{c}{0.00}
& \multicolumn{2}{c}{0.25}
& \multicolumn{2}{c}{0.50}
& \multicolumn{2}{c}{0.75}
& \multicolumn{2}{c}{1.00}
& \multicolumn{2}{c}{1.25}
& \multicolumn{2}{c}{1.50}
& \multicolumn{2}{c}{1.75}
& \multicolumn{2}{c}{2.00} \\

\cmidrule(lr){1-2}\cmidrule(lr){3-4}\cmidrule(lr){5-6}\cmidrule(lr){7-8}
\cmidrule(lr){9-10}\cmidrule(lr){11-12}\cmidrule(lr){13-14}
\cmidrule(lr){15-16}\cmidrule(lr){17-18}\cmidrule(lr){19-20}

\multirow{1}{*}{$\sigma_v$} &
\multirow{1}{*}{\textbf{Model}} & M & C & M & C & M & C & M & C & M & C & M & C & M & C & M & C & M & C \\

\midrule

% =================== sigma = 0.25 ===================
\multirow{6}{*}{0.25}

& RandSmooth
& 97 & 76 & 95 & 59 & 93 & 37 & 87 & 22 & 0 & 0 & 0 & 0 & 0 & 0 & 0 & 0 & 0 & 0 \\

& SWEEN
& 98 & 77 & 97 & 61 & 94 & 46 & 90 & 30 & 0 & 0 & 0 & 0 & 0 & 0 & 0 & 0 & 0 & 0 \\

& $\text{CEAR}^-$(RW)
& 98 & \textbf{79} & 97 & \textbf{65} & 95 & 47 & \textbf{93} & 31 & 0 & 0 & 0 & 0 & 0 & 0 & 0 & 0 & 0 & 0 \\

& $\text{CEAR}^-$(GM)
& \textbf{99} & \textbf{79} & \textbf{98} & \textbf{65} & \textbf{96} & \textbf{51} & \textbf{93} & 31 & 0 & 0 & 0 & 0 & 0 & 0 & 0 & 0 & 0 & 0 \\

& CEAR(RW)
& 98 & 76 & 97 & 63 & 93 & 49 & 90 & \textbf{36} & 0 & 0 & 0 & 0 & 0 & 0 & 0 & 0 & 0 & 0 \\

& CEAR(GM)
& 98 & 72 & 97 & 58 & 93 & 45 & 90 & 33 & 0 & 0 & 0 & 0 & 0 & 0 & 0 & 0 & 0 & 0 \\

\midrule

% =================== sigma = 0.50 ===================
\multirow{6}{*}{0.50}

& RandSmooth
& 96 & 64 & 95 & 55 & 92 & 38 & 86 & 28 & 77 & 15 & 66 & 9 & 39 & 5 & 18 & 2 & 0 & 0 \\

& SWEEN
& 97 & 68 & \textbf{97} & 56 & 93 & \textbf{44} & 86 & 34 & 84 & 20 & 72 & 14 & 50 & 8 & 20 & 5 & 0 & 0 \\

& $\text{CEAR}^-$(RW)
& 97 & 68 & 94 & 56 & 92 & 42 & 87 & 31 & 75 & 22 & 66 & 14 & 51 & 9 & 26 & 4 & 0 & 0 \\

& $\text{CEAR}^-$(GM)
& \textbf{98} & \textbf{70} & \textbf{97} & \textbf{58} & \textbf{94} & \textbf{44} & \textbf{89} & 31 & \textbf{85} & \textbf{25} & \textbf{72} & 16 & 55 & 11 & 25 & 6 & 0 & 0 \\

& CEAR(RW)
& 94 & 68 & 92 & 56 & 87 & \textbf{44} & 83 & \textbf{36} & 75 & \textbf{23} & 67 & 14 & \textbf{58} & 10 & \textbf{43} & \textbf{8} & 0 & 5 \\

& CEAR(GM)
& 96 & 63 & 93 & 53 & 90 & 39 & 84 & 29 & 75 & 17 & 61 & 10 & 44 & 8 & 25 & 0 & 0 & 0 \\

\midrule

% =================== sigma = 1.00 ===================
\multirow{6}{*}{1.00}

& RandSmooth
& 87 & 42 & 80 & 31 & 71 & 23 & 59 & 15 & 42 & 11 & 28 & 7 & 15 & 6 & 9 & 3 & 5 & 1 \\

& SWEEN
& 88 & 43 & 80 & 34 & 71 & 26 & 64 & 20 & 50 & 16 & 37 & 12 & 28 & 9 & 15 & 6 & 9 & 4 \\

& $\text{CEAR}^-$(RW)
& 87 & \textbf{47} & 79 & 35 & 69 & 26 & 62 & 20 & 58 & 15 & 49 & 12 & 32 & 8 & 22 & 6 & 15 & 4 \\

& $\text{CEAR}^-$(GM)
& \textbf{91} & \textbf{47} & 83 & 37 & \textbf{76} & 27 & 69 & 22 & 57 & 16 & 42 & 13 & 31 & 11 & 17 & 8 & 13 & 5 \\

& CEAR(RW)
& 91 & \textbf{47} & \textbf{87} & \textbf{37} & \textbf{76} & \textbf{30} & \textbf{69} & \textbf{25} & \textbf{60} & \textbf{20} & \textbf{50} & \textbf{18} & \textbf{38} & \textbf{15} & \textbf{34} & \textbf{10} & \textbf{20} & \textbf{8} \\

& CEAR(GM)
& 88 & 46 & 80 & 35 & 72 & 29 & 58 & 24 & 45 & 18 & 36 & 15 & 22 & 12 & 12 & 10 & 7 & 8 \\

\bottomrule
\end{tabular}
}
\label{tab:MNIST_CIFAR_Merged}
\end{table*}

\begin{table*}
    \centering
    \caption{Certified accuracy (\%) at varying radii and $\sigma_v$ on TinyImageNet.}
    \vspace{-2mm}
    \resizebox{0.75\textwidth}{!}{
    \begin{tabular}{c l c c c c c c c c c c}
        \toprule
         \multirow{3}{*}{$\sigma_v$} & \multirow{3}{*}{\textbf{Model}} & \multicolumn{9}{c}{\textbf{ Radius (R)}} \\
        \cmidrule(lr){3-11}
        % & \multicolumn{1}{c}{0.00}
        % & \multicolumn{1}{c}{0.1}
        % & \multicolumn{1}{c}{0.2}
        % & \multicolumn{1}{c}{0.3}
        % & \multicolumn{1}{c}{0.4}
        % & \multicolumn{1}{c}{0.5}
        % & \multicolumn{1}{c}{0.6}
        % & \multicolumn{1}{c}{0.7}
        % & \multicolumn{1}{c}{0.8} \\
         &  & 0.00 & 0.1 & 0.2 & 0.3 & 0.4 & 0.5 & 0.6 & 0.7 & 0.8  \\
         % \cmidrule{1-2}
        
        \midrule
        \multirow{6}{*}{0.125} 
        & RandSmooth & 38 & 28 & 19 & 13 & 8 & 0 & 0 & 0 & 0\\
        & SWEEN       & 45 & 33 & 24 & 18 & 13 & 0 & 0 & 0 & 0\\
        % & CEAR + MV    & 0 & 0 & 0 & 0 & 0 & 0 & 0 & 0 & 0  \\
        & CEAR$^-$(RW)      & 45 & 35 & 30 & 24 & 16 & 0 & 0 & 0 & 0\\
        & CEAR$^-$(GM)      & 46 & \textbf{36} & 29 & 25 & 16 & 0 & 0 & 0 & 0\\
        % & CEAR + VGA + MV    & 45 & 33 & 29 & 25 & 18 & 0 & 0 & 0 & 0  \\
        & CEAR(RW)       & \textbf{48} & 35 & \textbf{30} & \textbf{26} & \textbf{20} & 0 & 0 & 0 & 0\\
        & CEAR(GM)       & 46 & 36 & 29 & 25 & 17 & 0 & 0 & 0 & 0\\

        \midrule
        \multirow{6}{*}{0.25} 
        & RandSmooth & 28 & 25 & 18 & 13 & 7 & 9 & 4 & 3 & 3\\
        & SWEEN       & 30 & 23 & 17 & 14 & 10 & 9 & 6 & 5 & 5\\
        % & CEAR + MV      & 0 & 0 & 0 & 0 & 0 & 0 & 0 & 0 & 0\\
        & CEAR$^-$(RW)    & 37 & 33 & 29 & 23 & 20 & 16 & 13 & 10 & 8\\
        & CEAR$^-$(GM)      & 37 & 33 & 27 & 20 & 17 & 12 & 9 & 7 & 6\\
        % & CEAR + VGA + MV      & 0 & 0 & 0 & 0 & 0 & 0 & 0 & 0 & 0\\
        & CEAR(RW)       & \textbf{42} & \textbf{35} & \textbf{31} & \textbf{24} & \textbf{20} & \textbf{17} & \textbf{13} & \textbf{10} & \textbf{9}\\
        & CEAR(GM)       & 32 & 29 & 26 & 21 & 15 & 11 & 9 & 7 & 5\\

        \bottomrule
    \end{tabular}
    }
\label{TinyImgNet_Table_exp}
\end{table*}

\vspace{1mm}
\noindent\textbf{Experiment 1 (Certified Accuracy).}
We use certified accuracy (CA)~\citep{cohen2019certified} at given radius $L$ as a metric to measure the proportion of instances that are correctly classified by the smoothed classifier and are provably robust to any adversarial perturbation as, 
\begin{equation}
\mathrm{CA}
= \frac{1}{|\mathcal{X}_v|} \sum_{j \in \mathcal{X}_v} \mathbb{I}\{R_j \ge L\}\ ,    
\end{equation}
where $\mathbb{I}\{\cdot\}$ is the indicator function and $R_j$ is the robustness radius for each sample $j$ that is correctly classified.

Tables~\ref{tab:MNIST_CIFAR_Merged} and~\ref{TinyImgNet_Table_exp} report CA as a function of the perturbation radius $R$ and noise level $\sigma_v$ for RandSmooth, SWEEN, and CEAR variants, providing empirical support for our main theoretical claim that the optimal aggregation strategy depends critically on model confidence and perturbation budget of the smoothed classifier. Across MNIST and CIFAR10, we observe that in high confidence regions (small $R$), $\text{CEAR}^-$(GM) is optimal; it consistently yields the highest robust accuracy, achieving $99\%$ at $R=0$ on MNIST ($\sigma_v=0.25$) and $51\%$ at $R=0.5$ on CIFAR10 ($\sigma_v=0.25$), outperforming both RandSmooth and SWEEN. However, as $R$ increases and predictive uncertainty grows, this advantage vanishes, and fixed-weight aggregation (RW) with VGA becomes more reliable for low-confidence regimes: CEAR(RW) attains $58\%$ at $R=1.50$ on MNIST ($\sigma_v=0.50$) and $20\%$ at $R=1.0$ on CIFAR10 ($\sigma_v=1.00$), surpassing both baselines. This pattern is even stronger on TinyImageNet, where inherently lower model confidence renders adaptive weighting less effective and fixed RW consistently dominates (e.g., $26\%$ at $R=0.3$, $\sigma_v=0.125$). Further experimental results and interpretations on certified accuracy are reported in Appendix~\ref{append:a1}.

\begin{figure}[!t]
\centering
% ---------- MNIST ----------
\textbf{MNIST}\par\smallskip
\begin{subfigure}{0.22\textwidth}
\includegraphics[width=\linewidth]{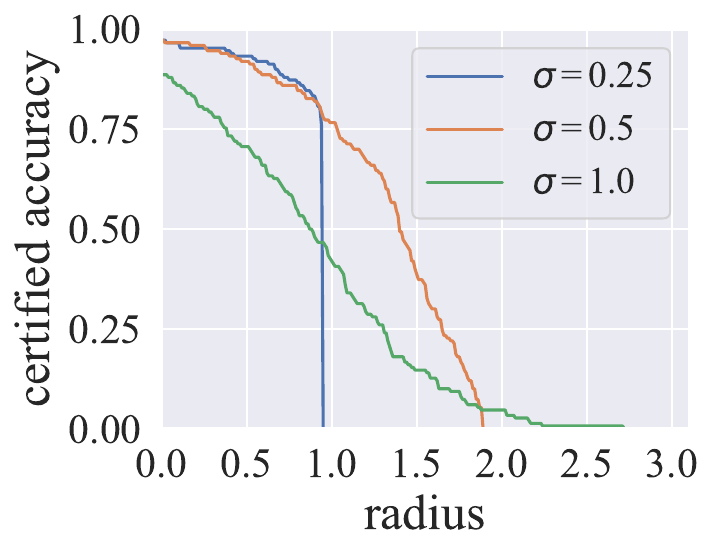}
\caption{RandSmooth}
\end{subfigure}
\begin{subfigure}{0.22\textwidth}
\includegraphics[width=\linewidth]{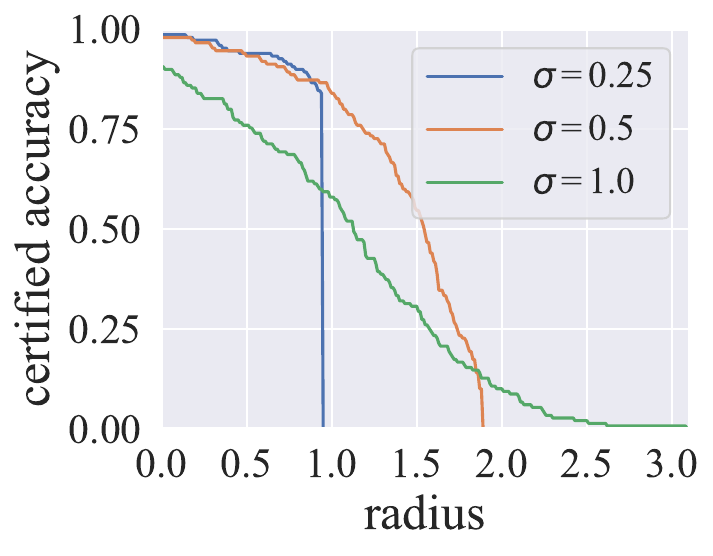}
\caption{SWEEN}
\end{subfigure}
\begin{subfigure}{0.22\textwidth}
\includegraphics[width=\linewidth]{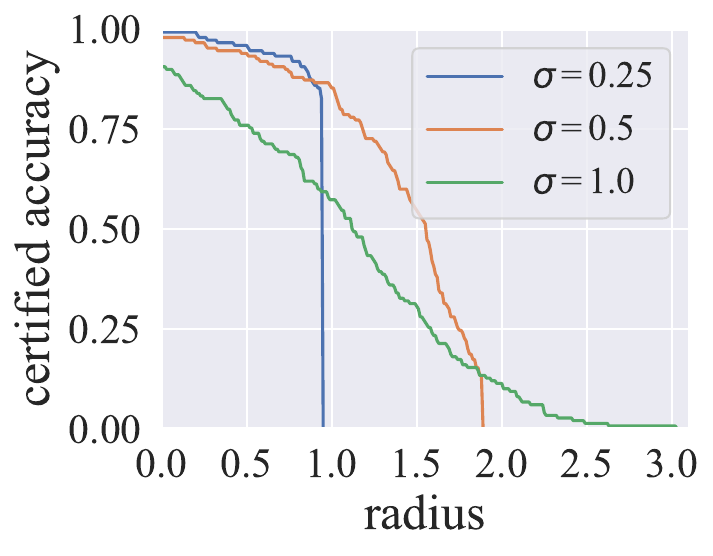}
\caption{$\text{CEAR}^-$(GM)}
\end{subfigure}
\begin{subfigure}{0.22\textwidth}
\includegraphics[width=\linewidth]{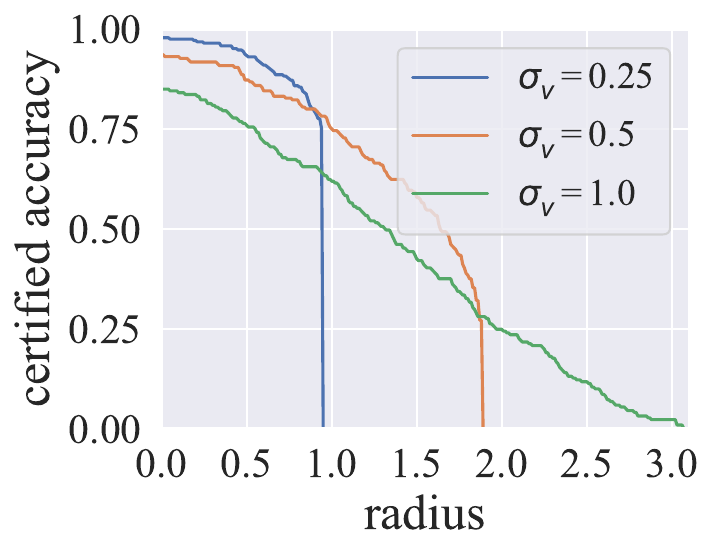}
\caption{CEAR(RW)}
\end{subfigure}
\vspace{2mm}

% ---------- CIFAR10 ----------
\textbf{CIFAR10}\par\smallskip
\begin{subfigure}{0.22\textwidth}
\includegraphics[width=\linewidth]{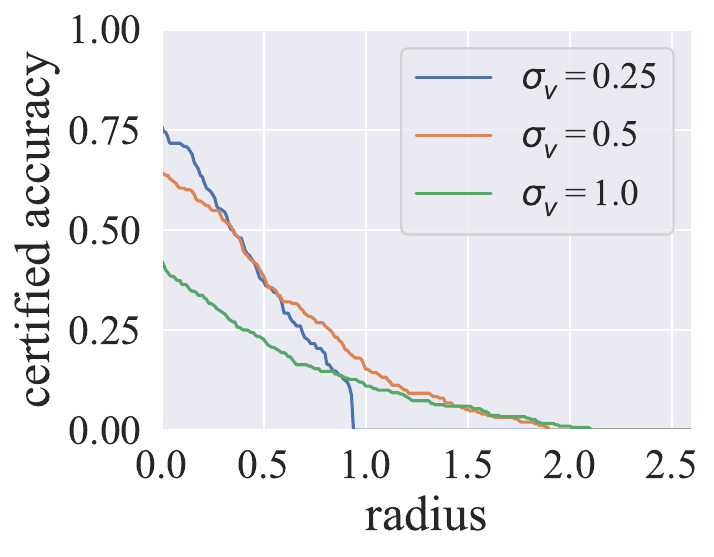}
\caption{RandSmooth}
\end{subfigure}
\begin{subfigure}{0.22\textwidth}
\includegraphics[width=\linewidth]{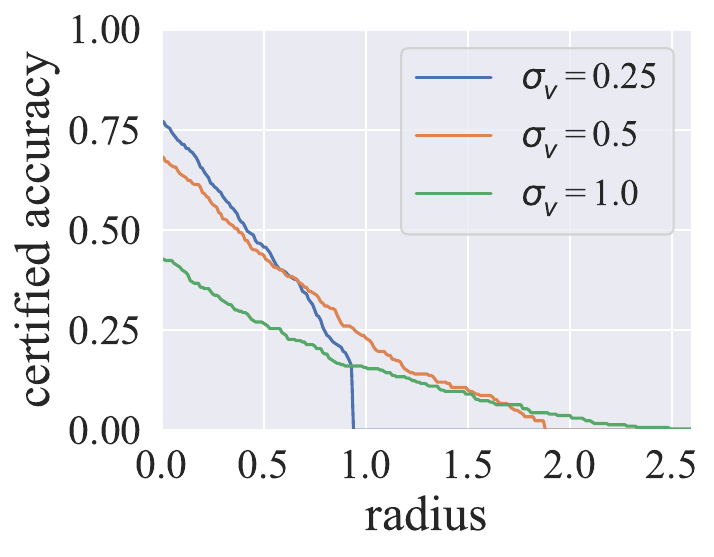}
\caption{SWEEN}
\end{subfigure}
\begin{subfigure}{0.22\textwidth}
\includegraphics[width=\linewidth]{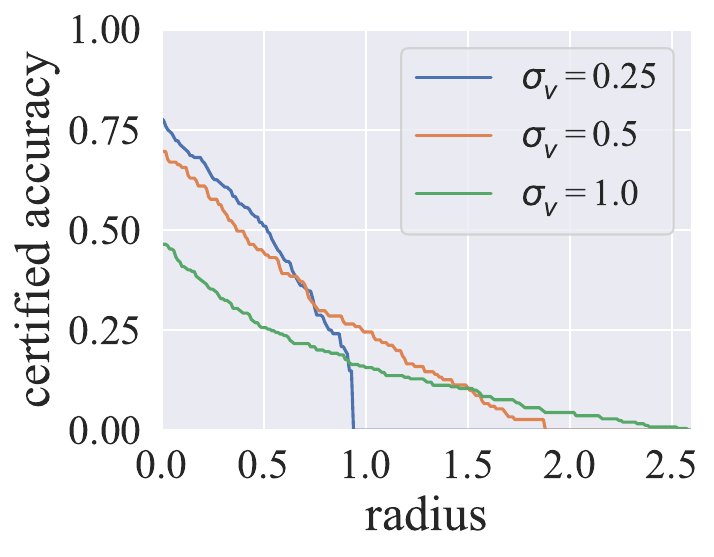}
\caption{$\text{CEAR}^-$(GM)}
\end{subfigure}
\begin{subfigure}{0.22\textwidth}
\includegraphics[width=\linewidth]{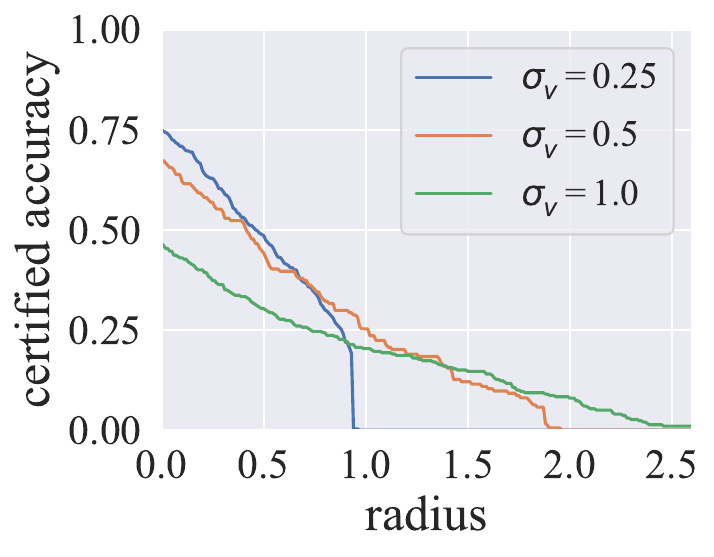}
\caption{CEAR(RW)}
\end{subfigure}
\vspace{2mm}

% ---------- TinyImageNet ----------
\textbf{TinyImageNet}\par%\smallskip
\begin{subfigure}{0.22\textwidth}
\includegraphics[width=\linewidth]{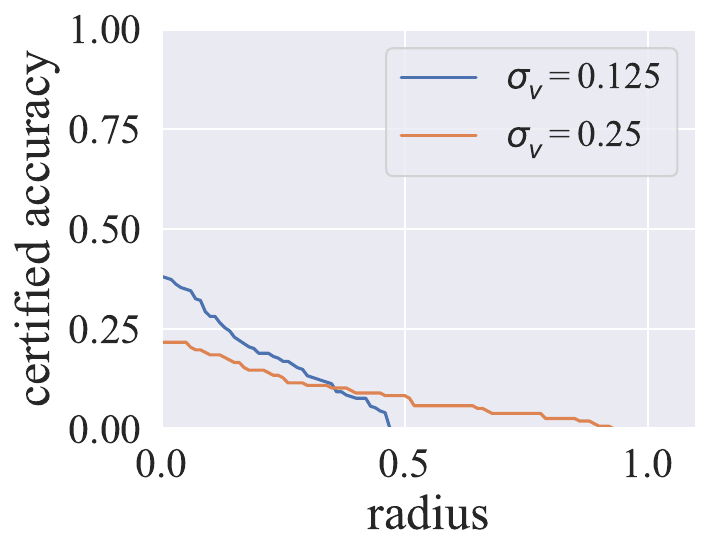}
\caption{RandSmooth}
\end{subfigure}
\begin{subfigure}{0.22\textwidth}
\includegraphics[width=\linewidth]{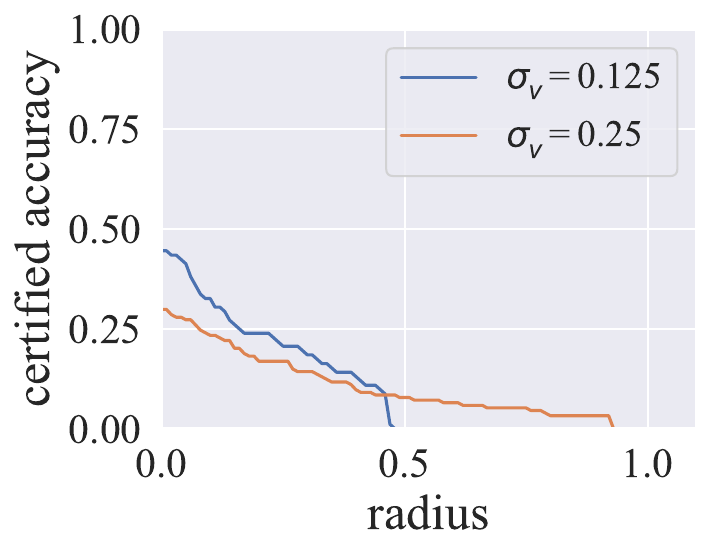}
\caption{SWEEN}
\end{subfigure}
\begin{subfigure}{0.22\textwidth}
\includegraphics[width=\linewidth]{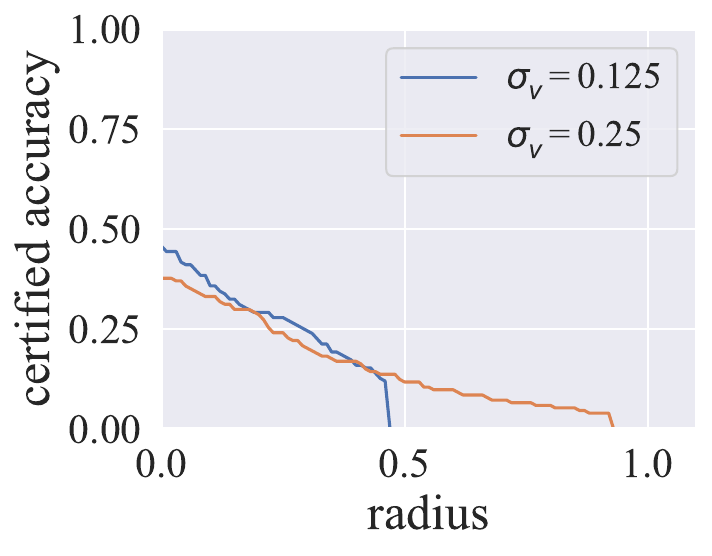}
\caption{$\text{CEAR}^-$(GM)}
\end{subfigure}
\begin{subfigure}{0.22\textwidth}
\includegraphics[width=\linewidth]{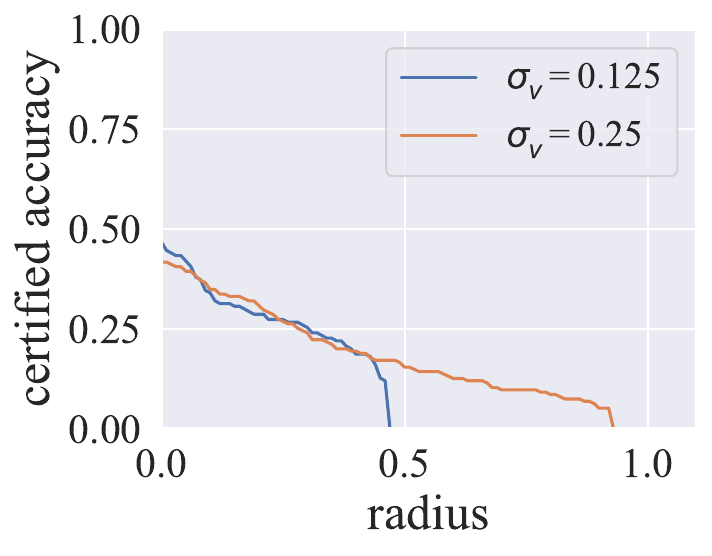}
\caption{CEAR(RW)}
\end{subfigure}
\caption{The average certified accuracy under varying radii and noise $\sigma_v$}
\label{fig:CertAllDatasets}
\vspace{-5mm}
\end{figure}

\vspace{1mm}
\noindent\textbf{Experiment 2 (Radius of Robustness).}
Figure~\ref{fig:CertAllDatasets} illustrates the certified robustness radius achieved by each method under increasing noise levels $\sigma_v$. Across all three datasets, CEAR(RW) consistently achieves a larger robustness radius than baselines and other voting mechanisms, indicating improved robustness in lower-confidence regions under higher perturbations. In particular, Figures~\ref{fig:CertAllDatasets}(d),~\ref{fig:CertAllDatasets}(h), and~\ref{fig:CertAllDatasets}(l) demonstrate that CEAR(RW) maintains a greater CA at $R > 1.0$ whereas RandSmooth and SWEEN rapidly deteriorate.

On MNIST in Figure~\ref{fig:CertAllDatasets}(c), CEAR(RW) exhibits a smoother decline in CA as the radius increases. For instance, at $\sigma_v=1.0$ and $R=2.5$, both RandSmooth and SWEEN fall to near-zero CA, whereas CEAR(RW) continues to dampen the effects of the perturbations and maintain its accuracy. A similar trend is observed on CIFAR10 and TinyImageNet in Figures~\ref{fig:CertAllDatasets}(h) and~\ref{fig:CertAllDatasets}(l), confirming that robust weighted aggregation is particularly effective in the large-radius regime. In contrast, Figure~\ref{fig:CertAllDatasets} shows that $\text{CEAR}^-$(GM) achieves the highest CA at small radii but degrades more rapidly beyond $R>1.0$. While this method still outperforms baselines at both lower and higher radii, it demonstrates that using a varied noise distribution in training in conjunction with an RW aggregation is better in low-confidence regions. Further experimental results and interpretations on radius of robustness are reported in Appendix~\ref{append:A2}. 

\vspace{1mm}
\noindent\textbf{Experiment 3 (Mitigating Transferability Impact). } 
We evaluated the transferability by generating attacks targeting one network at a time and measuring attacker success rate across the ensemble. Networks with low transferability achieve a lower attacker success rate. In contrast, to provide a non-transferable comparison, RandSmooth employs a single network where adversarial examples are generated and tested on the same network.
As shown in Table~\ref{perform_imagenetval_Linfan2}, for instance, on CIFAR10 under the $\ell_\infty$ norm ($\varepsilon=0.03$), CEAR(RW) substantially reduces the attacker success rate to $46\%$, outperforming RandSmooth ($96.56\%$) and SWEEN ($60.10\%$), demonstrating markedly improved robustness to transferable attacks.

%% file: Sections/Conclusion.tex
\section{Conclusion}
\label{conclusion}
CEAR is an ensemble-based robustness framework that integrates empirical and certified defense mechanisms across training and inference. During training, CEAR couples DTS and VGA to diversify ensemble members, smooth decision boundaries, and reduce adversarial transferability. At inference time, noisy logits are used to obfuscate gradients, while ensemble aggregation recovers clean accuracy and improves CA. We study two aggregation strategies: an optimized geometric median (GM) and a fixed robust weighted ensemble (RW). Experiments show that GM achieves higher certified accuracy in high-confidence, low-perturbation regions, whereas RW coupled with varying noise is more reliable under larger perturbation budgets and lower-confidence settings, yielding larger certified radii and improved robustness compared to prior ensemble baselines. These results indicate that optimized aggregation is preferable when ensemble predictions are confident, while fixed robust weighting is superior in high-uncertainty adversarial regimes. For future work, we plan to systematically investigate how the choice of temperature in DTS influences ensemble diversity, calibration, and certified robustness across different perturbation regimes.

\begin{table*}[t]
\centering
\caption{The attacker success rates under $\ell_2$ and $\ell_{\infty}$ norms}
\vspace{-2mm}
\resizebox{0.75\textwidth}{!}{
\begin{tabular}{l ccc ccc}
\toprule
\textbf{Model} & 
\multicolumn{3}{c}{\textbf{MNIST}} & 
\multicolumn{3}{c}{\textbf{CIFAR10}}\\

\cmidrule(lr){2-4} \cmidrule(lr){5-7}

& Clean Acc. 
& $\ell_2\ (\varepsilon = 1.00)$ 
& $\ell_{\infty}\ (\varepsilon = 0.20)$
& Clean Acc. 
& $\ell_2\ (\varepsilon = 0.35)$ 
& $\ell_{\infty}\ (\varepsilon = 0.03)$\\

\midrule
RandSmooth 
& 99.12\% & 12.20\% & 99.82\%
& 75.00\% & 33.18\% & 96.56\%\\

\addlinespace
SWEEN
& 99.16\% & 11.18\% & 56.28\%
& 80.20\% & 45.29\% & 60.10\%\\

\addlinespace
CEAR$^-$(GM)
& 99.10\% & \textbf{9.07\%} & 50.13\%
& 84.6\% & 19.9\% & 48.31\%\\

\addlinespace
CEAR(RW)
& 98.29\% & 10.60\% & \textbf{44.60\%}
& 81.00\% & \textbf{16.79\%} & \textbf{46.00\%}\\

\bottomrule
\end{tabular}
}
\label{perform_imagenetval_Linfan2}
\end{table*}

%% file: Sections/Appendix.tex
\clearpage

% \title{HEAR: Hybrid Ensemble Adversarial Robustness in Deep Neural Networks\\ (Supplementary Materials)}
% \maketitle
% \thispagestyle{empty}

% \vspace*{0.2in}

\appendix

\section{Experimental Results}

\subsection{Extended Certified Accuracy Analysis}
\label{append:a1}

Tables~\ref{MNIST_Table},~\ref{Cifar10_Table}, and~\ref{TinyImgNet_Table} provide a complete breakdown of certified accuracy (CA) across perturbation radii and smoothing noise levels for all evaluated aggregation strategies. While the main paper presented the primary trends, the full tables reveal a more nuanced interaction between confidence, smoothing strength, and ensemble aggregation. Across all datasets, two distinct certification regimes emerge: (i) a high-confidence regime at small radii where preserving consensus between ensemble members is most important, and (ii) a low-confidence regime at larger radii where maintaining robustness under uncertainty becomes dominant.

On MNIST (Table~\ref{MNIST_Table}), the strongest performance at low perturbation budgets consistently arises from geometric aggregation. At $\sigma_v=0.25$, $\text{CEAR}^{-}$(GM) achieves the highest certified accuracy across nearly all non-zero radii, reaching $99\%$ at $R=0$, $98\%$ at $R=0.25$, and $96\%$ at $R=0.50$, outperforming both RandSmooth and SWEEN. Similar behavior persists at $\sigma_v=0.50$, where CEAR$^{-}$(GM) remains optimal through $R=1.25$, achieving $85\%$ certified accuracy at $R=1.0$ compared to $77\%$ for RandSmooth and $84\%$ for SWEEN. These results suggest that when confidence remains concentrated, geometric median aggregation effectively suppresses unstable predictions while preserving the dominant consensus of the ensemble. However, this advantage gradually disappears as the perturbation radius increases. At $\sigma_v=1.00$, where smoothing introduces substantially greater uncertainty, CEAR(RW) becomes dominant, achieving $87\%$ at $R=0.25$, $76\%$ at $R=0.50$, $60\%$ at $R=1.00$, and maintaining $20\%$ certified accuracy at $R=2.00$, outperforming all alternative methods. This transition indicates that weighted aggregation becomes increasingly beneficial as confidence margins shrink and preserving predictions from stronger ensemble members becomes more important than enforcing geometric consensus.

A similar but more pronounced trend appears on CIFAR10 (Table~\ref{Cifar10_Table}). Compared with MNIST, certified accuracy deteriorates more rapidly as the perturbation radius increases, reflecting the higher input dimensionality and greater semantic complexity of the dataset. Under low smoothing noise ($\sigma_v=0.25$), CEAR$^{-}$(GM) achieves the strongest performance at moderate radii, reaching $51\%$ certified accuracy at $R=0.50$, compared to $46\%$ for SWEEN and $37\%$ for RandSmooth. However, unlike MNIST, geometric aggregation loses effectiveness earlier as uncertainty increases. At $\sigma_v=0.50$, CEAR(RW) begins to dominate from approximately $R=0.75$ onward, achieving $36\%$ at $R=0.75$ and maintaining non-trivial certification at larger radii where baseline methods collapse. This effect becomes strongest at $\sigma_v=1.00$, where CEAR(RW) consistently produces the highest certified accuracy across nearly the entire radius spectrum, achieving $30\%$ at $R=0.50$, $25\%$ at $R=0.75$, $20\%$ at $R=1.00$, and preserving $8\%$ certified accuracy even at $R=2.00$. The persistence of certification at large radii suggests that robust weighted aggregation benefits from assigning greater influence to ensemble members that remain reliable under heavy smoothing.

The behavior becomes even more evident on TinyImageNet (Table~\ref{TinyImgNet_Table}), where lower baseline confidence substantially alters the optimal aggregation strategy. Unlike MNIST and CIFAR10, robust weighted aggregation consistently dominates across almost all evaluated radii and smoothing configurations. At $\sigma_v=0.125$, CEAR(RW) achieves the highest certified accuracy at $R=0$ ($48\%$), maintains $30\%$ at $R=0.20$, and reaches $20\%$ at $R=0.40$, outperforming both SWEEN and geometric aggregation. This advantage becomes larger under $\sigma_v=0.25$, where CEAR(RW) achieves $42\%$ at $R=0$, $31\%$ at $R=0.20$, $24\%$ at $R=0.30$, and maintains non-zero certification through $R=0.80$. In contrast, geometric median aggregation degrades more rapidly as confidence decreases. These findings indicate that as dataset difficulty increases and classifier outputs become less concentrated, adaptive geometric consensus becomes less effective than explicitly weighting stronger ensemble members.
Tables~\ref{MNIST_Table}-\ref{TinyImgNet_Table} reinforce the central hypothesis of this work: there is no universally optimal aggregation strategy for certified robustness. Instead, the optimal mechanism depends on the confidence regime induced by randomized smoothing. Geometric aggregation is most effective when ensemble predictions remain concentrated and stable, whereas robust weighted aggregation becomes increasingly advantageous as uncertainty grows. This transition becomes more pronounced as dataset complexity increases, ultimately explaining the strong performance of CEAR(RW) on large-scale certification tasks.

\begin{table*}
    \centering
    \caption{Certified accuracy at varying radii and $\sigma_v$ on MNIST} 
    \vspace{-2mm}
    \resizebox{0.9\textwidth}{!}{%
    \begin{tabular}{c l c c c c c c c c c c}
        \toprule
        \multirow{0}{*}{$\sigma_v$} & \multirow{0}{*}{\textbf{Model}} & \multicolumn{9}{c}{\textbf{Certified accuracy (\%) on different radii (R)}}  \\
        \cmidrule(lr){3-11}
         & & 0.00 & 0.25 & 0.50 & 0.75 & 1.00 & 1.25 & 1.50 & 1.75 & 2.00  \\
        \midrule
        \multirow{6}{*}{0.25} 
        & RandSmooth & 97 & 95 & 93 & 87 & 0 & 0 & 0 & 0 & 0\\
        & SWEEN       & 98 & 97 & 94 & 90 & 0 & 0 & 0 & 0 & 0\\
        & CEAR$^-$(MV)       & 98 & 97 & 95 & 92 & 0 & 0 & 0 & 0 & 0\\
        & CEAR$^-$(RW)      & 98 & 97 & 95 & \textbf{93} & 0 & 0 & 0 & 0 & 0\\
        & CEAR$^-$(GM)    & \textbf{99} & \textbf{98} & \textbf{96} & \textbf{93} & 0 & 0 & 0 & 0 & 0  \\
        & CEAR(MV)       & 98 & 95 & 93 & 89 & 0 & 0 & 0 & 0 & 0\\
        & CEAR(RW)    & 98 & 97 & 93 & 90 & 0 & 0 & 0 & 0 & 0  \\
        & CEAR(GM)    & 98 & 97 & 93 & 90 & 0 & 0 & 0 & 0 & 0  \\

        \midrule
        \multirow{8}{*}{0.50} 
        & RandSmooth & 96 & 95 & 92 & 86 & 77 & 66 & 39 & 18 & 0\\
        & SWEEN       & 97 & \textbf{97} & 93 & 86 & 84 & 72 & 50 & 20 & 0\\
        & CEAR$^-$(MV)       & 96 & 93 & 90 & 84 & 78 & 70 & 51 & 28 & 0\\
        & CEAR$^-$(RW)   & 97 & 94 & 92 & 87 & 75 & 66 & 51 & 26 & 0  \\
        & CEAR$^-$(GM)       & \textbf{98} & \textbf{97} & \textbf{94} & \textbf{89} & \textbf{85} & \textbf{72} & 55 & 25 & 0\\
        & CEAR(MV)       & 95 & 95 & 93 & 83 & 78 & 70 & 56 & 35 & 0\\
        & CEAR(RW)       & 94 & 92 & 87 & 83 & 75 & 67 & \textbf{58} & \textbf{43} & 0\\
        & CEAR(GM)    & 96 & 93 & 90 & 84 & 75 & 61 & 44 & 25 & 0  \\
        
        \midrule
        \multirow{7}{*}{1.00} 
        & RandSmooth & 87 & 80 & 71 & 59 & 42 & 28 & 15 & 9 & 5\\
        & SWEEN       & 88 & 80 & 71 & 64 & 50 & 37 & \textbf{28} & 15 & 9\\
        & CEAR$^-$(MV)       & 83 & 80 & 70 & 64 & 52 & 41 & 30 & 18 & 11\\
        & CEAR$^-$(RW)   & 87 & 79 & 69 & 62 & 58 & 49 & 32 & 22 & 15  \\
        & CEAR$^-$(GM)   & \textbf{91} & 83 & \textbf{76} & 69 & 57 & 42 & 31 & 17 & 13\\
        & CEAR(MV)       & 87 & 81 & 73 & 66 & 54 & 41 & 32 & 26 & 12\\
        & CEAR(RW)       & 91 & \textbf{87} & 76 & \textbf{69} & \textbf{60} & \textbf{50} & 38 & \textbf{34} & \textbf{20}\\
        & CEAR(GM)    & 88 & 80 & 72 & 58 & 45 & 36 & 22 & 12 & 7  \\
        \bottomrule
    \end{tabular}%
    }
    \label{MNIST_Table}
\end{table*}

\begin{table*}
    \centering
    \caption{Certified accuracy at varying radii and $\sigma_v$ on CIFAR10}
    \vspace{-2mm}
    \resizebox{0.9\textwidth}{!}{%
    \begin{tabular}{c l c c c c c c c c c c}
        \toprule
        \multirow{0}{*}{$\sigma_v$} & \multirow{0}{*}{\textbf{Model}} & \multicolumn{9}{c}{\textbf{Certified accuracy (\%) on different radii (R)}} \\
        \cmidrule(lr){3-11}
         &  & 0.00 & 0.25 & 0.50 & 0.75 & 1.00 & 1.25 & 1.50 & 1.75 & 2.00  \\
        \midrule
        \multirow{8}{*}{0.25} 
        & RandSmooth & 76 & 59 & 37 & 22 & 0 & 0 & 0 & 0 & 0\\
        & SWEEN       & 77 & 61 & 46 & 30 & 0 & 0 & 0 & 0 & 0\\
        & CEAR$^-$(MV)    & 72 & 58 & 44 & 32 & 0 & 0 & 0 & 0 & 0  \\
        & CEAR$^-$(RW)      & \textbf{79} & \textbf{65} & 47 & 31 & 0 & 0 & 0 & 0 & 0\\
        & CEAR$^-$(GM)    & 79 & 65 & \textbf{51} & 31 & 0 & 0 & 0 & 0 & 0 \\
        & CEAR(MV)    & 72 & 60 & 46 & 35 & 0 & 0 & 0 & 0 & 0  \\
        & CEAR(RW)    & 76 & 63 & 49 & \textbf{36} & 0 & 0 & 0 & 0 & 0  \\
        & CEAR(GM)    & 72 & 58 & 45 & 33 & 0 & 0 & 0 & 0 & 0  \\

        \midrule
        \multirow{8}{*}{0.50} 
        & RandSmooth        & 64 & 55 & 38 & 28 & 15 & 9 & 5 & 2 & 0\\
        & SWEEN             & 68 & 56 & \textbf{44} & 34 & 20 & 14 & 8 & 5 & 0\\
        & CEAR$^-$(MV)         & 61 & 50 & 38 & 32 & 23 & 16 & 10 & 7 & 0\\
        & CEAR$^-$(RW)        & 68 & 56 & 42 & 31 & 22 & 14 & 9 & 4 & 0 \\
        & CEAR$^-$(GM)         & \textbf{70} & \textbf{58} & \textbf{44} & 31 & 25 & 16 & 11 & 6 & 0 \\
        & CEAR(MV)   & 63 & 55 & 41 & 33 & \textbf{24} & \textbf{17} & \textbf{11} & 7 & 0\\
        & CEAR(RW)  & \textbf{68} & 56 & \textbf{44} & \textbf{36} & 23 & 14 & 10 & \textbf{8} & 5\\
        & CEAR(GM)   & 63 & 53 & 39 & 29 & 17 & 10 & 8 & 0 & 0  \\
        
        \midrule
        \multirow{8}{*}{1.00} 
        & RandSmooth & 42 & 31 & 23 & 15 & 11 & 7 & 6 & 3 & 1\\
        & SWEEN       & 43 & 34 & 26 & 20 & 16 & 12 & 9 & 6 & 4\\
        & CEAR$^-$(MV)    & 42 & 33 & 26 & 20 & 16 & 15 & 10 & 6 & 5  \\
        & CEAR$^-$(RW)    & \textbf{47} & 35 & 26 & 20 & 15 & 12 & 8 & 6 & 4  \\
        & CEAR$^-$(GM)    & 47 & 37 & 27 & 22 & 16 & 13 & 11 & 8 & 5 \\
        & CEAR(MV)    & 43 & 34 & 26 & 21 & 17 & 15 & 12 & 7 & 5  \\
        & CEAR(RW)       & \textbf{47} & \textbf{37} & \textbf{30} & \textbf{25} & \textbf{20} & \textbf{18} & \textbf{15} & \textbf{10} & \textbf{8}\\
        & CEAR(GM)    & 46 & 35 & 29 & 24 & 18 & 15 & 12 & 10 & 8  \\

        \bottomrule
    \end{tabular}%
    }
\label{Cifar10_Table}
\end{table*}

\begin{table*}
    \centering
    \caption{Certified accuracy (\%) at varying radii and $\sigma_v$ on TinyImageNet.}
    \vspace{-2mm}
    \resizebox{0.75\textwidth}{!}{
    \begin{tabular}{c l c c c c c c c c c c}
        \toprule
         \multirow{3}{*}{$\sigma_v$} & \multirow{3}{*}{\textbf{Model}} & \multicolumn{9}{c}{\textbf{ Radius (R)}} \\
        \cmidrule(lr){3-11}
        % & \multicolumn{1}{c}{0.00}
        % & \multicolumn{1}{c}{0.1}
        % & \multicolumn{1}{c}{0.2}
        % & \multicolumn{1}{c}{0.3}
        % & \multicolumn{1}{c}{0.4}
        % & \multicolumn{1}{c}{0.5}
        % & \multicolumn{1}{c}{0.6}
        % & \multicolumn{1}{c}{0.7}
        % & \multicolumn{1}{c}{0.8} \\
         &  & 0.00 & 0.1 & 0.2 & 0.3 & 0.4 & 0.5 & 0.6 & 0.7 & 0.8  \\
         % \cmidrule{1-2}
        
        \midrule
        \multirow{8}{*}{0.125} 
        & RandSmooth & 38 & 28 & 19 & 13 & 8 & 0 & 0 & 0 & 0\\
        & SWEEN       & 45 & 33 & 24 & 18 & 13 & 0 & 0 & 0 & 0\\
        & CEAR$^-$(MV)    & 48 & \textbf{37} & \textbf{31} & 25 & 19 & 0 & 0 & 0 & 0  \\
        & CEAR$^-$(RW)      & 45 & 35 & 30 & 24 & 16 & 0 & 0 & 0 & 0\\
        & CEAR$^-$(GM)      & 46 & \textbf{36} & 29 & 25 & 16 & 0 & 0 & 0 & 0\\
        & CEAR(MV)    & 45 & 33 & 29 & 25 & 18 & 0 & 0 & 0 & 0  \\
        & CEAR(RW)       & \textbf{48} & 35 & \textbf{31} & \textbf{26} & \textbf{20} & 0 & 0 & 0 & 0\\
        & CEAR(GM)       & 46 & 36 & 29 & 25 & 17 & 0 & 0 & 0 & 0\\

        \midrule
        \multirow{8}{*}{0.25} 
        & RandSmooth & 28 & 25 & 18 & 13 & 7 & 9 & 4 & 3 & 3\\
        & SWEEN       & 30 & 23 & 17 & 14 & 10 & 9 & 6 & 5 & 5\\
        & CEAR$^-$(MV)      & 40 & 32 & 28 & 20 & 17 & 16 & 10 & 9 & 5\\
        & CEAR$^-$(RW)    & 37 & 33 & 29 & 23 & 20 & 16 & 12 & 9 & 8\\
        & CEAR$^-$(GM)      & 37 & 33 & 27 & 20 & 17 & 12 & 9 & 7 & 6\\
        & CEAR(MV)      & 37 & 33 & 27 & 22 & 19 & 14 & 12 & 8 & 7\\
        & CEAR(RW)       & \textbf{42} & \textbf{35} & \textbf{31} & \textbf{24} & \textbf{20} & \textbf{17} & \textbf{13} & \textbf{10} & \textbf{9}\\
        & CEAR(GM)       & 32 & 29 & 26 & 21 & 15 & 11 & 9 & 7 & 5\\

        \bottomrule
    \end{tabular}
    }
\label{TinyImgNet_Table}
\end{table*}

\subsection{Extended Certified Accuracy Curves Across Smoothing Levels}
\label{append:A2}
\begin{figure*}[t]
\centering

% Row 1
\begin{subfigure}{0.32\textwidth}
\centering
\includegraphics[width=\linewidth]{Sections/Images/Cohen_MNIST.pdf}
\caption{RandSmooth}
\label{RSMnist}
\end{subfigure}
\hfill
\begin{subfigure}{0.32\textwidth}
\centering
\includegraphics[width=\linewidth]{Sections/Images/SWEEN_MNIST.pdf}
\caption{SWEEN}
\label{SWEENMnist}
\end{subfigure}
\hfill
\begin{subfigure}{0.32\textwidth}
\centering
\includegraphics[width=\linewidth]{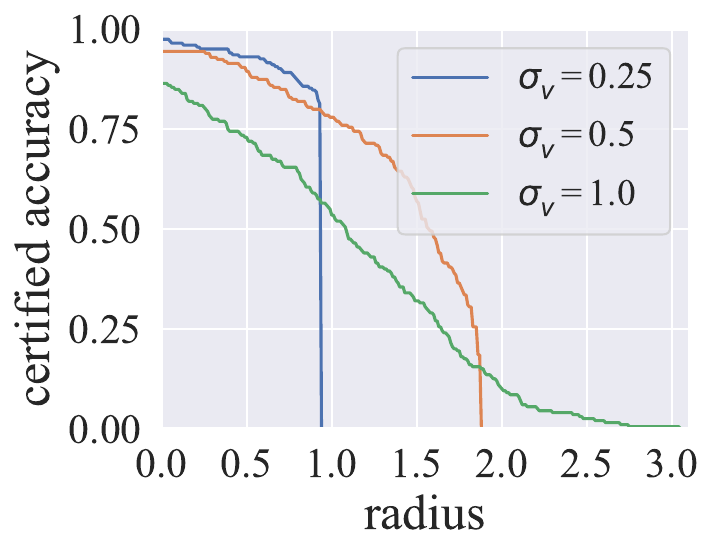}
\caption{CEAR$^-$(MV)}
\label{CEARMVMnist}
\end{subfigure}

\vspace{0.5em}

% Row 2
\begin{subfigure}{0.32\textwidth}
\centering
\includegraphics[width=\linewidth]{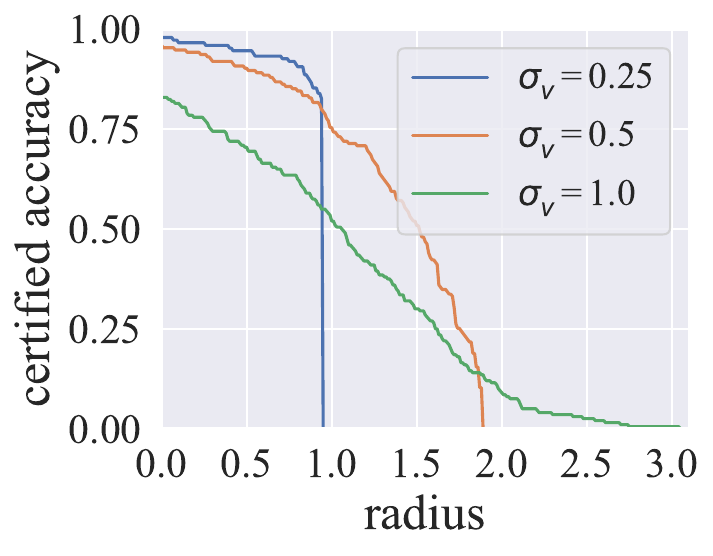}
\caption{CEAR$^-$(RW)}
\label{CEARWEMnist}
\end{subfigure}
\hfill
\begin{subfigure}{0.32\textwidth}
\centering
\includegraphics[width=\linewidth]{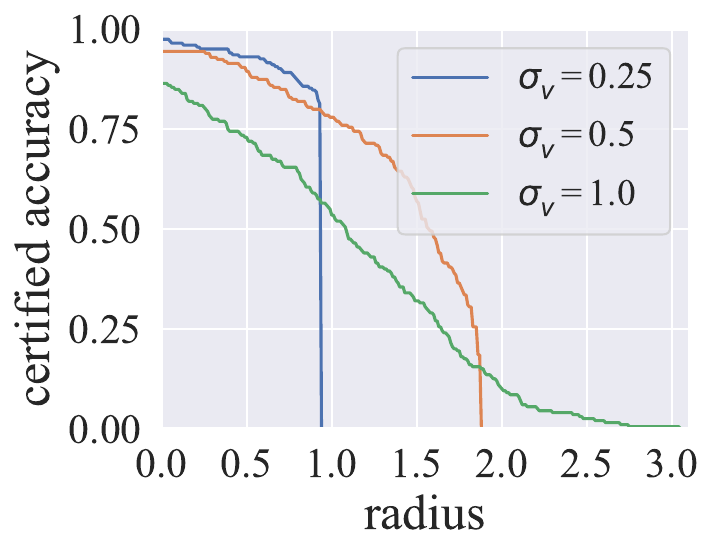}
\caption{CEAR(MV)}
\label{CEARVGAMVMnist}
\end{subfigure}
\hfill
\begin{subfigure}{0.32\textwidth}
\centering
\includegraphics[width=\linewidth]{Sections/Images/Proposed_WE_WVGA_MNIST.pdf}
\caption{CEAR(RW)}
\label{CEARVGAWEMnist}
\end{subfigure}

\vspace{0.5em}

% Row 3 (centered)
\makebox[\textwidth][c]{
\begin{subfigure}{0.32\textwidth}
\centering
\includegraphics[width=\linewidth]{Sections/Images/proposed_GM_WOVGA_MNIST.pdf}
\caption{CEAR$^-$(GM)}
\label{CEARGMMnist}
\end{subfigure}
\hspace{0.03\textwidth}
\begin{subfigure}{0.32\textwidth}
\centering
\includegraphics[width=\linewidth]{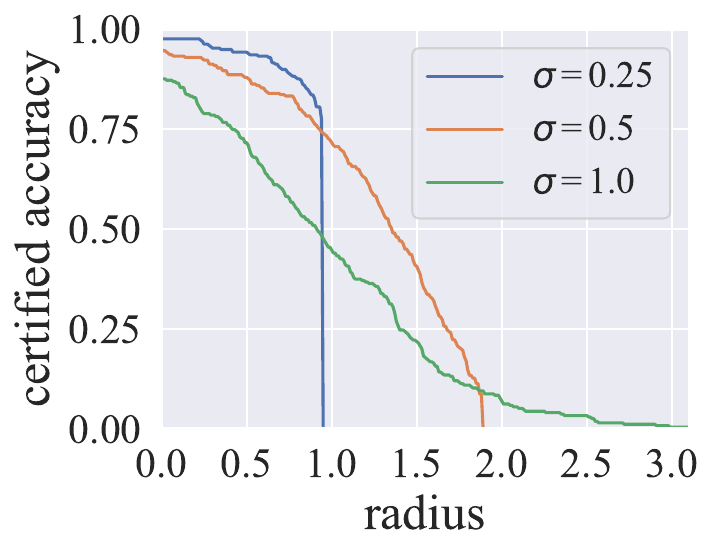}
\caption{CEAR(GM)}
\label{CEARVGAGMMnist}
\end{subfigure}
}
\caption{The certified accuracy for MNIST under varying radii and $\sigma_v$}
\label{fig:CertRadiusMNIST}

\end{figure*}

% ---- this is for Cifar10 ---

\begin{figure*}[t]
\centering

% Row 1
\begin{subfigure}{0.32\textwidth}
\centering
\includegraphics[width=\linewidth]{Sections/Images/Cohen_Cifar10.pdf}
\caption{RandSmooth}
\label{RSCifar10}
\end{subfigure}
\hfill
\begin{subfigure}{0.32\textwidth}
\centering
\includegraphics[width=\linewidth]{Sections/Images/SWEEN_Cifar10.pdf}
\caption{SWEEN}
\label{SWEENCifar10}
\end{subfigure}
\hfill
\begin{subfigure}{0.32\textwidth}
\centering
\includegraphics[width=\linewidth]{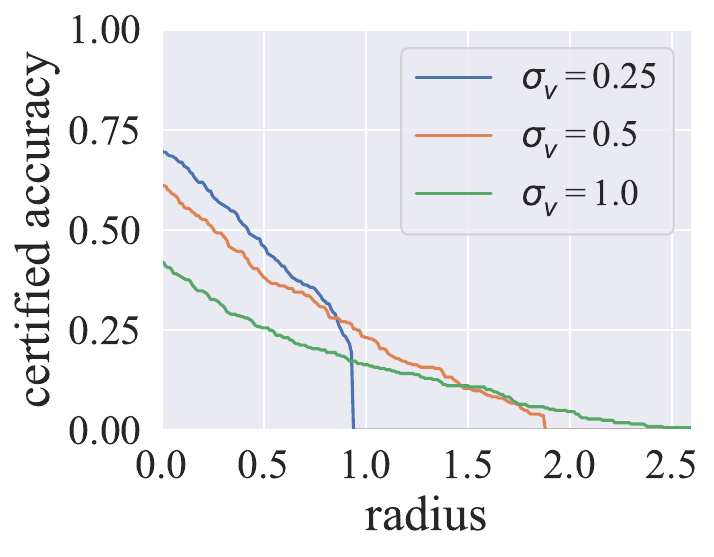}
\caption{CEAR$^-$(MV)}
\label{CEARMVCifar10}
\end{subfigure}

\vspace{0.5em}

% Row 2
\begin{subfigure}{0.32\textwidth}
\centering
\includegraphics[width=\linewidth]{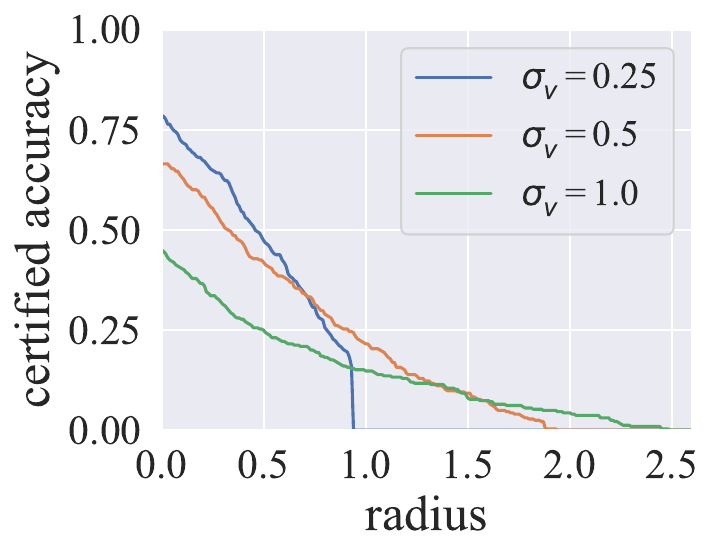}
\caption{CEAR$^-$(RW)}
\label{CEARWECifar10}
\end{subfigure}
\hfill
\begin{subfigure}{0.32\textwidth}
\centering
\includegraphics[width=\linewidth]{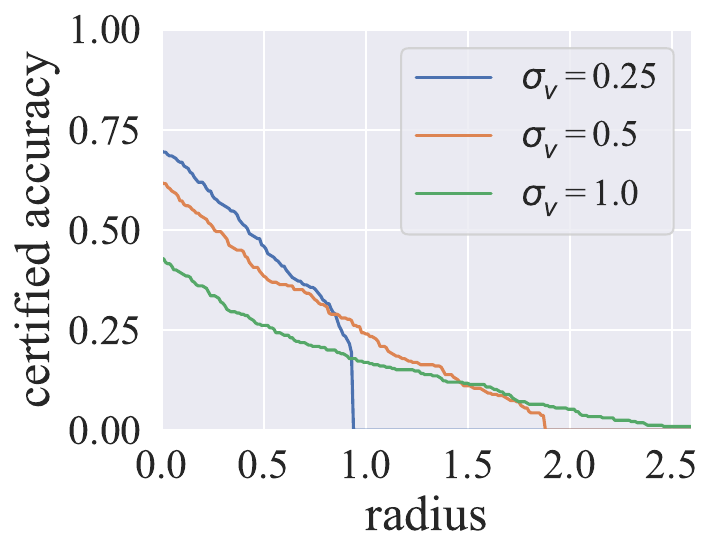}
\caption{CEAR(MV)}
\label{CEARVGAMVCifar10}
\end{subfigure}
\hfill
\begin{subfigure}{0.32\textwidth}
\centering
\includegraphics[width=\linewidth]{Sections/Images/Proposed_VGA_WE_Cifar10.pdf}
\caption{CEAR(RW)}
\label{CEARVGAWECifar10}
\end{subfigure}

\vspace{0.5em}

% Row 3 (centered)
\makebox[\textwidth][c]{
\begin{subfigure}{0.32\textwidth}
\centering
\includegraphics[width=\linewidth]{Sections/Images/Proposed_GM_WOVGA_Cifar10.pdf}
\caption{CEAR$^-$(GM)}
\label{CEARGMCifar10}
\end{subfigure}
\hspace{0.03\textwidth}
\begin{subfigure}{0.32\textwidth}
\centering
\includegraphics[width=\linewidth]{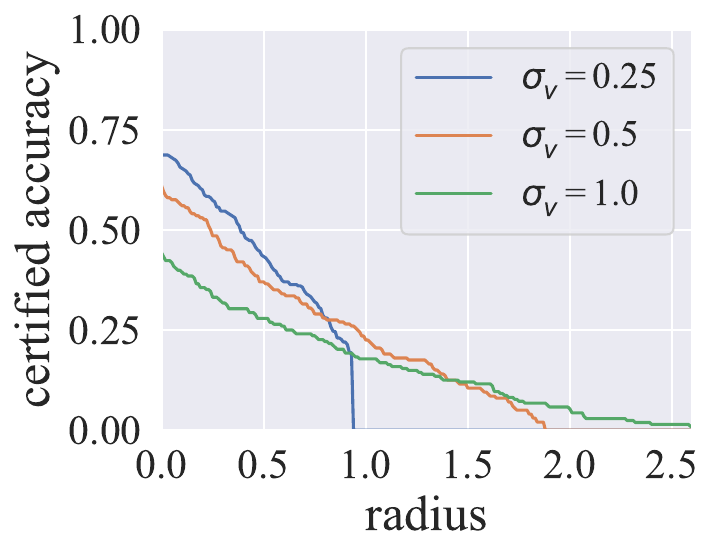}
\caption{CEAR(GM)}
\label{CEARVGAGMCifar10}
\end{subfigure}
}

\caption{The certified accuracy for CIFAR-10 under varying radii and $\sigma_v$}
\label{fig:CertRadiusCifar10}

\end{figure*}

% % ---- Tiny-ImageNet Certified Accuracy ----
% \begin{figure*}[t]
% \centering

% % Row 1
% \begin{subfigure}{0.32\textwidth}
% \centering
% \includegraphics[width=\linewidth]{Sections/Images/Cohen_TinyImgNet.pdf}
% \caption{RandSmooth}
% \label{RSTinyImgNet}
% \end{subfigure}
% \hfill
% \begin{subfigure}{0.32\textwidth}
% \centering
% \includegraphics[width=\linewidth]{Sections/Images/SWEEN_TinImgNet.pdf}
% \caption{SWEEN}
% \label{SWEENTinyImgNet}
% \end{subfigure}
% \hfill
% \begin{subfigure}{0.32\textwidth}
% \centering
% \includegraphics[width=\linewidth]{Sections/Images/TinImgNet_Proposed_WE_WOVGA.pdf}
% \caption{CEAR$^-$(RW)}
% \label{CEARWETinyImgNet}
% \end{subfigure}

% \vspace{0.5em}

% % Row 2
% \begin{subfigure}{0.32\textwidth}
% \centering
% \includegraphics[width=\linewidth]{Sections/Images/Proposed_WE_WVGA_TinyImgNet.pdf}
% \caption{CEAR(RW)}
% \label{CEARVGAWETinyImgNet}
% \end{subfigure}
% \hfill
% \begin{subfigure}{0.32\textwidth}
% \centering
% \includegraphics[width=\linewidth]{Sections/Images/Proposed_GM_WOVGA_TinImgNet.pdf}
% \caption{CEAR$^-$(GM)}
% \label{CEARGMTinyImgNet}
% \end{subfigure}
% \hfill
% \begin{subfigure}{0.32\textwidth}
% \centering
% \includegraphics[width=\linewidth]{Sections/Images/TinImgNet_Proposed_GM_WVGA.pdf}
% \caption{CEAR(GM)}
% \label{CEARVGAGMTinyImgNet}
% \end{subfigure}

% \caption{Average certified accuracy for Tiny-ImageNet under varying radii and noise $\sigma_v$.}
% \label{fig:CertRadiusTinyImgNet}

% \end{figure*}

% ---- This is the images for TinyImgNet ----
\begin{figure*}[t]
\centering

% Row 1
\begin{subfigure}{0.32\textwidth}
\centering
\includegraphics[width=\linewidth]{Sections/Images/Cohen_TinyImgNet.pdf}
\caption{RandSmooth}
\label{RSTinyImgNet}
\end{subfigure}
\hfill
\begin{subfigure}{0.32\textwidth}
\centering
\includegraphics[width=\linewidth]{Sections/Images/SWEEN_TinImgNet.pdf}
\caption{SWEEN}
\label{SWEENTinyImgNet}
\end{subfigure}
\hfill
\begin{subfigure}{0.32\textwidth}
\centering
\includegraphics[width=\linewidth]{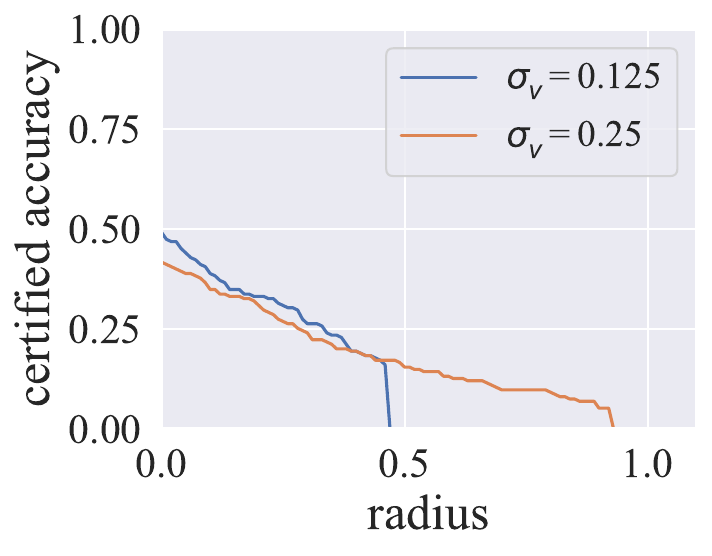}
\caption{CEAR$^-$(MV)}
\label{CEARMVTinyImgNet}
\end{subfigure}

\vspace{0.5em}

% Row 2
\begin{subfigure}{0.32\textwidth}
\centering
\includegraphics[width=\linewidth]{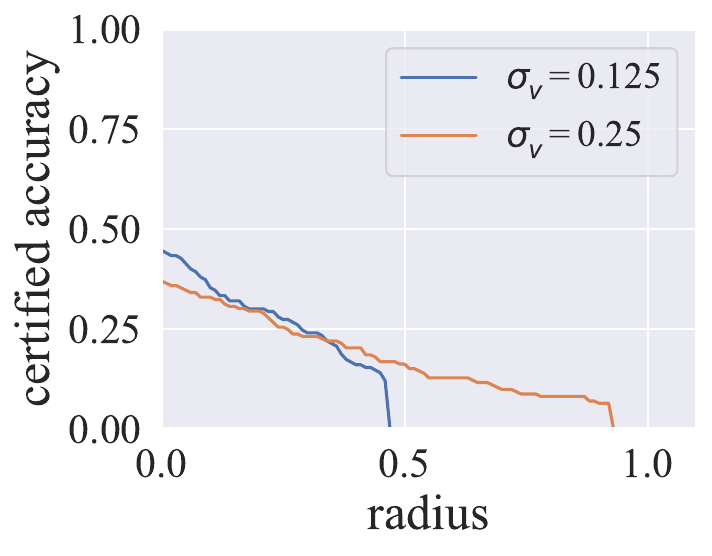}
\caption{CEAR$^-$(RW)}
\label{CEARWETinyImgNet}
\end{subfigure}
\hfill
\begin{subfigure}{0.32\textwidth}
\centering
\includegraphics[width=\linewidth]{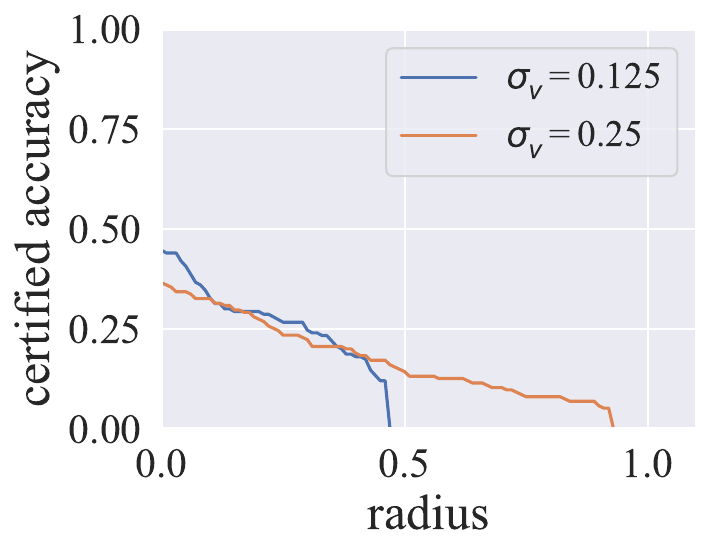}
\caption{CEAR(MV)}
\label{CEARVGAMVTinyImgNet}
\end{subfigure}
\hfill
\begin{subfigure}{0.32\textwidth}
\centering
\includegraphics[width=\linewidth]{Sections/Images/Proposed_WE_WVGA_TinyImgNet.pdf}
\caption{CEAR(RW)}
\label{CEARVGAWETinyImgNet}
\end{subfigure}

\vspace{0.5em}

% Row 3 (centered)
\makebox[\textwidth][c]{
\begin{subfigure}{0.32\textwidth}
\centering
\includegraphics[width=\linewidth]{Sections/Images/Proposed_GM_WOVGA_TinImgNet.pdf}
\caption{CEAR$^-$(GM)}
\label{CEARGMTinyImgNet}
\end{subfigure}
\hspace{0.03\textwidth}
\begin{subfigure}{0.32\textwidth}
\centering
\includegraphics[width=\linewidth]{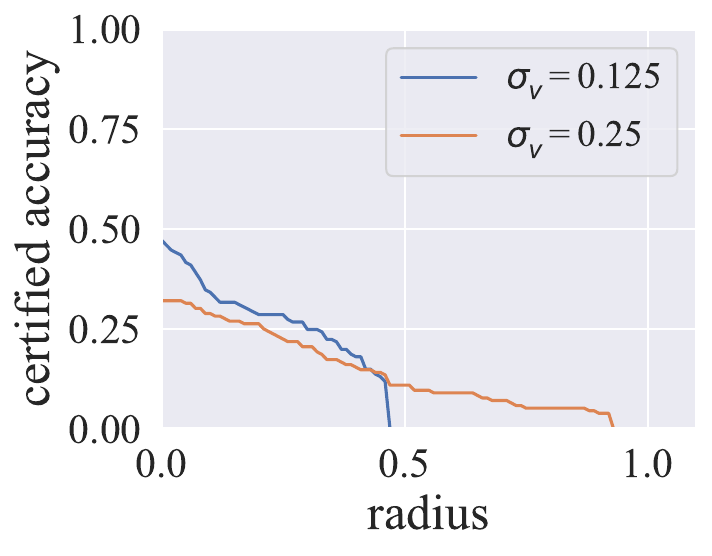}
\caption{CEAR(GM)}
\label{CEARVGAGMTinyImgNet}
\end{subfigure}
}

\caption{The certified accuracy for TinyImageNet under varying radii and $\sigma_v$}
\label{fig:CertRadiusTinyImgNet}

\end{figure*}

Figures~\ref{fig:CertRadiusMNIST}, \ref{fig:CertRadiusCifar10}, and \ref{fig:CertRadiusTinyImgNet} present the certified accuracy curves for all evaluated methods under varying smoothing noise levels $\sigma_v$. These curves provide a more detailed view of how certified accuracy degrades as the perturbation radius increases and complement the summarized results presented in Tables~\ref{MNIST_Table}--\ref{TinyImgNet_Table}. Consistent with the results reported in the main body of this paper, all methods exhibit a monotonic decrease in certified accuracy as the certification radius increases. Furthermore, increasing the smoothing variance generally shifts robustness toward larger perturbation budgets, resulting in a slower decline in certified accuracy at the cost of reduced performance in high-confidence regions.

A notable observation across all three datasets is that models trained with Variable Gaussian Augmentation (VGA) exhibit a slower degradation in certified accuracy than their non-VGA counterparts. This effect is most pronounced on MNIST when comparing CEAR(RW) in Figure~\ref{fig:CertRadiusMNIST}(f) with CEAR$^{-}$(RW) in Figure~\ref{fig:CertRadiusMNIST}(d). Examining the $\sigma_v=1.0$ curves (green), CEAR$^{-}$(RW) rapidly approaches zero certified accuracy once $R > 2.0$, whereas CEAR(RW) maintains approximately $25\%$ certified accuracy beyond the same radius. Since MNIST is a comparatively simple classification task, the benefits of VGA are visually apparent in the certification curves. Although the effect is less obvious on the more complex CIFAR10 and TinyImageNet datasets due to their lower baseline certified accuracies, the same trend remains present. This observation is further supported by Tables~\ref{MNIST_Table}--\ref{TinyImgNet_Table}, where VGA-based models consistently achieve higher certified accuracies at moderate and large certification radii. Collectively, these results suggest that VGA increases ensemble diversity and reduces correlated prediction collapse under randomized smoothing, allowing the ensemble to retain stronger robustness guarantees as prediction uncertainty increases.

%% file: references.bib
@inproceedings{lecuyer2019certified,
  title={Certified robustness to adversarial examples with differential privacy},
  author={Lecuyer, Mathias and Atlidakis, Vaggelis and Geambasu, Roxana and Hsu, Daniel and Jana, Suman},
  booktitle={symp. on security and privacy (SP)},
  pages={656--672},
  year={2019},
  organization={IEEE}
}

@inproceedings{papernot2016distillation,
  title={Distillation as a defense to adversarial perturbations against deep neural networks},
  author={Papernot, Nicolas and McDaniel, Patrick and Wu, Xi and Jha, Somesh and Swami, Ananthram},
  booktitle={symp. on security and privacy},
  pages={582--597},
  year={2016},
  organization={IEEE}
}

@inproceedings{yazdani2024denl,
  title={DENL: Diverse Ensemble and Noisy Logits for Improved Robustness of Neural Networks},
  author={Yazdani, Mina and Karimi, Hamed and Samavi, Reza},
  booktitle={ACML},
  pages={1574--1589},
  year={2024},
  organization={PMLR}
}

@inproceedings{carlini2017towards,
  title={Towards evaluating the robustness of neural networks},
  author={Carlini, Nicholas and Wagner, David},
  booktitle={2017 ieee symposium on security and privacy (sp)},
  pages={39--57},
  year={2017},
  organization={Ieee}
}

@article{liang2023advanced,
  title={Advanced defensive distillation with ensemble voting and noisy logits},
  author={Liang, Yuting and Samavi, Reza},
  journal={Applied Intelligence},
  volume={53},
  number={3},
  pages={3069--3094},
  year={2023},
  publisher={Springer}
}

@inproceedings{cohen2019certified,
  title={Certified adversarial robustness via randomized smoothing},
  author={Cohen, Jeremy and Rosenfeld, Elan and Kolter, Zico},
  booktitle={international conference on machine learning},
  pages={1310--1320},
  year={2019},
  organization={PMLR}
}

@inproceedings{NEURIPS2019_3a24b25a,
 author = {Salman, Hadi and Li, Jerry and Razenshteyn, Ilya and Zhang, Pengchuan and Zhang, Huan and Bubeck, Sebastien and Yang, Greg},
 booktitle = {NeurIPS},
 editor = {H. Wallach and H. Larochelle and A. Beygelzimer and F. d\textquotesingle Alch\'{e}-Buc and E. Fox and R. Garnett},
 publisher = {Curran Associates, Inc.},
 title = {Provably Robust Deep Learning via Adversarially Trained Smoothed Classifiers},
 volume = {32},
 year = {2019}
}

@misc{LeCun_Burges_Cortes_2010,
  author       = {Yann LeCun and Corinna Cortes and Christopher J. C. Burges},
  title        = {The MNIST Database of Handwritten Digits},
  year         = {1998},
  howpublished = {\url{http://yann.lecun.com/exdb/mnist/}}
}

@misc{Krizhevsky_2012,
  author       = {Alex Krizhevsky},
  title        = {Learning Multiple Layers of Features from Tiny Images},
  year         = {2009},
  howpublished = {Technical Report, University of Toronto},
  note         = {CIFAR-10 dataset},
  url          = {https://www.cs.toronto.edu/~kriz/learning-features-2009-TR.pdf}
}

@inproceedings{he2016deep,
  title={Deep residual learning for image recognition},
  author={He, Kaiming and Zhang, Xiangyu and Ren, Shaoqing and Sun, Jian},
  booktitle={Proc. of the IEEE conf. on computer vision and pattern recognition},
  pages={770--778},
  year={2016}
}

@inproceedings{yangcertified,
  title={On the Certified Robustness for Ensemble Models and Beyond},
  author={Yang, Zhuolin and Li, Linyi and Xu, Xiaojun and Kailkhura, Bhavya and Xie, Tao and Li, Bo},
  booktitle={Int. Conf. on Learning Representations},
    year={2022}
}

@inproceedings{liu2021enhancing,
  title={Enhancing Certified Robustness via Smoothed Weighted Ensembling},
  author={Liu, Chizhou and Feng, Yunzhen and Wang, Ranran and Dong, Bin},
  booktitle={ICML 2021 Workshop on Adv. Machine Learning},
    year={2020}
}

@inproceedings{croce2020reliable,
  title={Reliable evaluation of adversarial robustness with an ensemble of diverse parameter-free attacks},
  author={Croce, Francesco and Hein, Matthias},
  booktitle={ICML},
  pages={2206--2216},
  year={2020},
  organization={PMLR}
}

@inproceedings{szegedy2014intriguing,
  title={Intriguing properties of neural networks},
  author={Szegedy, Christian and Zaremba, Wojciech and Sutskever, Ilya and Bruna, Joan and Erhan, Dumitru and Goodfellow, Ian and Fergus, Rob},
  booktitle={2nd International Conference on Learning Representations, ICLR 2014},
  year={2014}
}

@inproceedings{jimenez1998dynamically,
  title={Dynamically weighted ensemble neural networks for classification},
  author={Jim{\'e}nez, Daniel},
  booktitle={1998 IEEE International Joint Conference on Neural Networks Proceedings. IEEE World Congress on Computational Intelligence (Cat. No. 98CH36227)},
  volume={1},
  pages={753--756},
  year={1998},
  organization={IEEE}
}

@inproceedings{papernot2016limitations,
  title={The limitations of deep learning in adversarial settings},
  author={Papernot, Nicolas and McDaniel, Patrick and Jha, Somesh and Fredrikson, Matthew and Celik, Z Berkay and Swami, Ananthram},
  booktitle={2016 IEEE European Symposium on Security and Privacy (EuroS\&P)},
  pages={372--387},
  year={2016},
  organization={IEEE}
}

@inproceedings{guo2017calibration,
  title={On calibration of modern neural networks},
  author={Guo, Chuan and Pleiss, Geoff and Sun, Yu and Weinberger, Kilian Q},
  booktitle={International Conference on Machine Learning},
  pages={1321--1330},
  year={2017},
  organization={PMLR}
}

@inproceedings{abadi2016tensorflow,
  title={TensorFlow: a system for Large-Scale machine learning},
  author={Abadi, Mart{\'\i}n and Barham, Paul and Chen, Jianmin and Chen, Zhifeng and Davis, Andy and Dean, Jeffrey and Devin, Matthieu and Ghemawat, Sanjay and Irving, Geoffrey and Isard, Michael and others},
  booktitle={12th USENIX symposium on OSDI 16},
  pages={265--283},
  year={2016}
}

@inproceedings{zhai2020macer,
  title={MACER: attack-free and scalable robust training via maximizing certified radius},
  author={Zhai, Runtian and Dan, Chen and He, Di and Zhang, Huan and Gong, Boqing and Ravikumar, Pradeep and Hsieh, Cho-Jui and Wang, Liwei},
  booktitle={International Conference on Learning Representations (ICLR)},
  year={2020}
}

@inproceedings{zhang2021towards,
  title={Towards certified robustness under label noise},
  author={Zhang, Kaidi and Zhu, Hongyang and Li, Xuechen and Evans, David},
  booktitle={International Conference on Learning Representations (ICLR)},
  year={2021}
}

@inproceedings{li2022double,
  title={Double-boosted randomized smoothing: A sharp certified defense against adversarial attacks},
  author={Li, Bo and Wang, Yihan and Ren, Zhi and Kolter, J. Zico},
  booktitle={NeurIPS},
  year={2022}
}

@inproceedings{li2023sok,
  title={Sok: Certified robustness for deep neural networks},
  author={Li, Linyi and Xie, Tao and Li, Bo},
  booktitle={2023 IEEE symposium on security and privacy (SP)},
  pages={1289--1310},
  year={2023},
  organization={IEEE}
}

@article{raghunathan2018semidefinite,
  title={Semidefinite relaxations for certifying robustness to adversarial examples},
  author={Raghunathan, Aditi and Steinhardt, Jacob and Liang, Percy S},
  journal={NeurIPS},
  volume={31},
  year={2018}
}

@article{lecun2002gradient,
  title={Gradient-based learning applied to document recognition},
  author={LeCun, Yann and Bottou, L{\'e}on and Bengio, Yoshua and Haffner, Patrick},
  journal={Proceedings of the IEEE},
  volume={86},
  number={11},
  pages={2278--2324},
  year={2002},
  publisher={Ieee}
}

@inproceedings{athalye2018obfuscated,
  title={Obfuscated gradients give a false sense of security: Circumventing defenses to adversarial examples},
  author={Athalye, Anish and Carlini, Nicholas and Wagner, David},
  booktitle={International Conference on Machine Learning},
  pages={274--283},
  year={2018},
  organization={PMLR}
}

@article{aldahdooh2022adversarial,
  title={Adversarial example detection for DNN models: A review and experimental comparison},
  author={Aldahdooh, Ahmed and Hamidouche, Wassim and Fezza, Sid Ahmed and D{\'e}forges, Olivier},
  journal={Artificial Intelligence Review},
  volume={55},
  number={6},
  pages={4403--4462},
  year={2022},
  publisher={Springer}
}

@article{chen2024diversity,
  title={Diversity supporting robustness: Enhancing adversarial robustness via differentiated ensemble predictions},
  author={Chen, Xi and Huang, Wei and Peng, Ziwen and Guo, Wei and Zhang, Fan},
  journal={Computers \& Security},
  volume={142},
  pages={103861},
  year={2024},
  publisher={Elsevier}
}

@article{huang2023fasten,
  title={FASTEN: fast ensemble learning for improved adversarial robustness},
  author={Huang, Lifeng and Huang, Qiong and Qiu, Peichao and Wei, Shuxin and Gao, Chengying},
  journal={IEEE Transactions on Information Forensics and Security},
  volume={19},
  pages={2565--2580},
  year={2023},
  publisher={IEEE}
}

@article{zuhlke2025adversarial,
  title={Adversarial robustness of neural networks from the perspective of lipschitz calculus: A survey},
  author={Z{\"u}hlke, Monty and Kudenko, Daniel},
  journal={ACM Computing Surveys},
  volume={57},
  number={6},
  pages={1--41},
  year={2025},
  publisher={ACM New York, NY}
}

@article{wang2025failure,
  title={Failure Cases Are Better Learned But Boundary Says Sorry: Facilitating Smooth Perception Change for Accuracy-Robustness Trade-Off in Adversarial Training},
  author={Wang, Yanyun and Liu, Li},
  journal={arXiv preprint arXiv:2508.02186},
  year={2025}
}

@inproceedings{deng2009imagenet,
  title     = {ImageNet: A Large-Scale Hierarchical Image Database},
  author    = {Deng, Jia and Dong, Wei and Socher, Richard and Li, Li-Jia and Li, Kai and Fei-Fei, Li},
  booktitle = {Proceedings of the IEEE Conference on Computer Vision and Pattern Recognition (CVPR)},
  year      = {2009},
  pages     = {248--255}
}

@article{haldane1948note,
  title={Note on the median of a multivariate distribution},
  author={Haldane, JBS},
  journal={Biometrika},
  volume={35},
  number={3-4},
  pages={414--417},
  year={1948},
  publisher={Oxford University Press}
}

@article{weiszfeld1937point,
  title={Sur le point pour lequel la somme des distances de n points donn{\'e}s est minimum},
  author={Weiszfeld, Endre},
  journal={Tohoku Mathematical Journal, First Series},
  volume={43},
  pages={355--386},
  year={1937},
  publisher={Mathematical Institute, Tohoku University}
}
